\documentclass[runningheads]{llncs}

\usepackage{eccv}

\usepackage{eccvabbrv}

\usepackage{graphicx}
\usepackage{booktabs}
\usepackage{algorithm}
\usepackage{algpseudocode}
\usepackage{multirow}
\usepackage{multicol}
\usepackage{flushend}
\usepackage{balance}

\usepackage[accsupp]{axessibility}  % Improves PDF readability for those with disabilities.

\usepackage{hyperref}
\usepackage{xcolor}

\usepackage{orcidlink}

\begin{document}

\title{Driving is a Game: Combining Planning and Prediction 
with Bayesian Iterative Best Response} 

\titlerunning{BIBeR: Bayesian Iterative Best Response}

\author{
\normalsize
Aron Distelzweig$^{1*}$, 
Yiwei Wang$^{2,3*}$, 
Faris Janjo\v{s}$^{2}$,
Marcel Hallgarten$^{2}$, \\
Mihai Dobre$^{4}$,
\normalsize
Alexander Langmann$^{3}$, 
Joschka Boedecker$^{1,5}$,
Johannes Betz$^{3}$\\[4pt]
\footnotesize}

\authorrunning{A. Distelzweig et al.}

\institute{$^{1}$ University of Freiburg, Germany \\ $^{2}$ Bosch Center for Artificial Intelligence, Germany \quad \\  $^{3}$ Technical University of Munich, Germany \\  $^{4}$ Klaris AI, UK \quad $^{5}$   BrainLinks-BrainTools, Germany\\
\footnotesize
$^{*}$ These authors contributed equally to this work.}

\maketitle

\begin{abstract}
Autonomous driving planning systems perform nearly perfectly in routine scenarios using lightweight, rule-based methods but still struggle in dense urban traffic, where lane changes and merges require anticipating and influencing other agents. Modern motion predictors offer highly accurate forecasts, yet their integration into planning is mostly rudimental: discarding unsafe plans. Similarly, end-to-end models offer a one-way integration that avoids the challenges of joint prediction and planning modeling under uncertainty. In contrast, game-theoretic formulations offer a principled alternative but have seen limited adoption in autonomous driving. We present Bayesian Iterative Best Response (BIBeR), a framework that unifies motion prediction and game-theoretic planning into a single interaction-aware process. BIBeR is the first to integrate a state-of-the-art predictor into an Iterative Best Response (IBR) loop, repeatedly refining the strategies of the ego vehicle and surrounding agents. This repeated best-response process approximates a Nash equilibrium, enabling bidirectional adaptation where the ego both reacts to and shapes the behavior of others. In addition, our proposed Bayesian confidence estimation quantifies prediction reliability and modulates update strength—more conservative under low confidence and more decisive under high confidence. BIBeR is compatible with modern predictors and planners, combining the transparency of structured planning with the flexibility of learned models. Experiments show that BIBeR achieves an 11\% improvement over state-of-the-art planners on highly interactive interPlan lane-change scenarios, while also outperforming existing approaches on standard nuPlan benchmarks.
\end{abstract}
\section{Introduction}
\begin{figure*}
    \includegraphics[width=\textwidth]{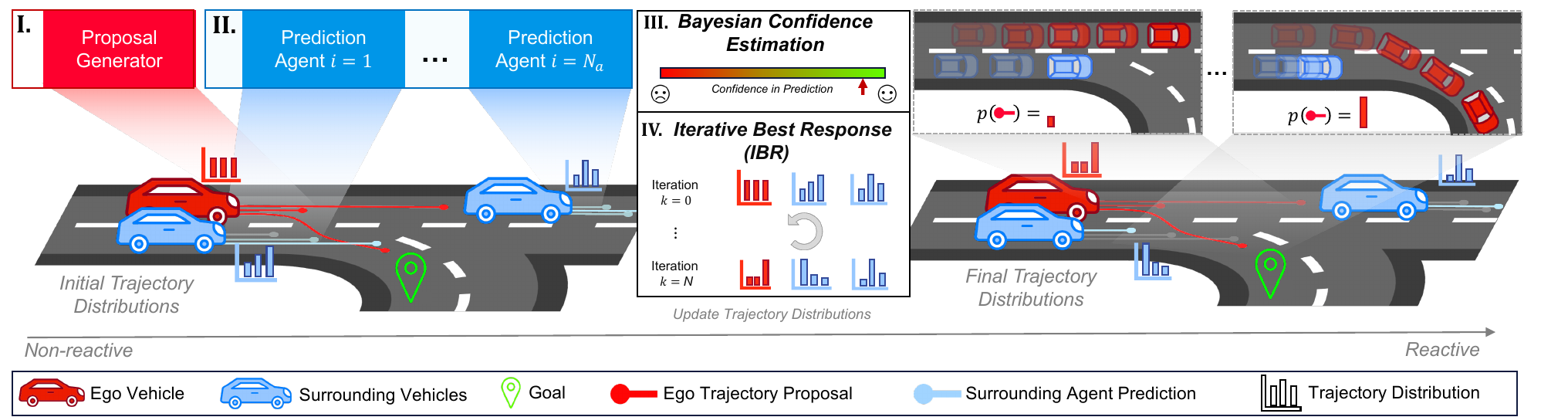}
    \caption{Overview of the proposed \textbf{Bayesian Iterative Best Response (BIBeR)} framework. (\textit{i}) the framework first generates a set of candidate trajectories for the ego vehicle, where the proposals are uniformly distributed. (\textit{ii}) prediction model forecasts the independent future motions of surrounding agents. (\textit{iii}) Bayesian confidence estimation balances assertiveness and conservatism in BIBeR: for each non-ego agent, a confidence term measures how well the updated trajectory matches observed motion versus the original prediction, directly guiding the weight update. (\textit{iv}) Iterative Best Response (IBR) procedure refines these predictions: each agent, including the ego, iteratively re-weights its fixed set of candidate trajectories to optimally respond to the others, thereby approximating a game-theoretic equilibrium.}
    \vspace{-0.4cm}
    \label{fig:biber}
\end{figure*}
Autonomous driving planning systems have made remarkable progress in recent years. State-of-the-art methods can already handle the majority of routine driving scenarios with high accuracy, often relying on lightweight, rule-based approaches~\cite{dauner2023parting}, that dominate standard planning benchmarks~\cite{caesar2022nuplan}. However, while these methods perform well in structured settings, benchmarks have only recently begun to consider interactive driving scenarios~\cite{hallgarten2024interplan}—a domain in which these approaches exhibit significant limitations.
Interactive scenarios require more than passive reproduction of typical driving behavior. In these settings, a driving system must not only perceive the current state of surrounding traffic but also infer the intentions of other agents and anticipate their reactions to the ego vehicle’s actions. For example, a driver executing a lane change may deliberately nudge towards an adjacent lane to induce a trailing vehicle to yield, a behavior that cannot be captured by non-reactive planning alone. Effectively leveraging this knowledge is essential for safe and efficient planning. 

Predicting the future behavior of traffic participants is a central component of this challenge and has been extensively studied under the task of motion prediction.
Motion prediction has advanced rapidly, with numerous models achieving state-of-the-art performance in forecasting agent trajectories~\cite{kim2021_lapred, liu2024laformer, deo2022_pgp, janjovs2023conditional, janjoš2022starnet, janjoš2023bridginggapmultisteponeshot, nayakanti2023_wayformer, jiang2023motiondiffuser, hallgarten2023prediction, yuan2021agentformer, gilles2022THOMAS, gilles2022GOHOME, ngiam2022scenetransformer}. Yet, this predictive power has not consistently translated into robust decision-making in interactive scenarios. Modern successful planning approaches~\cite{dauner2023parting, distelzweig2025perfectprediction} evaluate large sets of motion proposals, generated in a sampling-based manner, using learned or hand-crafted cost functions. While effective, these approaches treat predictions primarily as filters and do not reason about the deliberate interactions between agents.

An alternative paradigm comes from game-theoretic methods, which explicitly model interactions between agents. Iterative best-response (IBR) techniques, for instance, represent the decision-making of multiple drivers by repeatedly inferring each agent’s optimal strategy, assuming the others’ strategies are fixed. This approach provides an intuitive mechanism to balance competing interests in multi-agent settings. However, recent game-theoretic planners~\cite{huang2023gameformer, wang2021GameTheoretic} have generally not taken advantage of the predictive power of modern motion forecasting methods.
%\newpage
In this work, we present \textbf{Bayesian Iterative Best Response (BIBeR)}, a framework that unifies the strengths of prediction and game-theoretic planning. The framework leverages state-of-the-art motion predictors to initialize an IBR process, enabling interaction-aware and interpretable planning. Available strategies of each agent are represented by the marginal distribution estimated from a motion predictor, while the ego strategy is generated using a rule-based proposal generator. Importantly, all components—predictors and proposal generators—are modular and can be easily exchanged, providing flexibility and compatibility with existing methods. We further introduce Bayesian confidence estimation, which quantifies the reliability of predicted behaviors and modulates the strength of iterative updates—promoting conservative adjustments under low confidence and more decisive responses when confidence is high.
Overall, BIBeR is: (\textit{i}) \textit{flexible}, as it integrates seamlessly with state-of-the-art predictors and planners, using their marginal, interaction-unaware distributions as an initial condition for computing an interaction-aware joint distribution; (\textit{ii}) \textit{optimal}, because its game-theoretic integration converges to an optimal weight distribution given the cost function;  (\textit{iii}) \textit{adaptable}, since it provides a mechanism to balance assertiveness and conservatism in planning decisions, and (\textit{iv}) \textit{principled}, its explicit modeling of bidirectional interaction between the ego agent and surrounding traffic participants captures how ego actions influence others and how others in turn influence ego. Empirically, these qualities produce a game-theoretic planner that outperforms existing methods not only on interactive and out-of-distribution driving benchmarks but also on routine driving scenarios.

In summary, we present three key contributions:
\begin{itemize}
\item We introduce a framework for interaction-aware planning that integrates state-of-the-art learning-based predictors with sampling-based motion planners through an Iterative Best Response (IBR) formulation.
\item We propose a Bayesian confidence estimation mechanism that regulates update dynamics and improves safety in interactive decision-making.
\item We outperform existing approaches by a large margin in challenging interactive driving scenarios, while also achieving consistent improvements in routine driving scenarios.
\end{itemize}

\section{Related Work}
\subsection{Prediction}
Accurately forecasting the motion of traffic participants is a core challenge in autonomous driving. The key difficulty is modeling future uncertainty in the context of complex interactions between road users and static context. Prediction models can be broadly categorized into marginal and joint.

Marginal predictors~\cite{gao2020vectornet, liu2024laformer, deo2022_pgp, kim2021_lapred, yuan2021agentformer, nayakanti2023_wayformer} generate a trajectory forecast for each agent independently by modeling the marginal distribution of future motion, effectively averaging over the possible behaviors of all other agents rather than conditioning on them. Thus, marginal predictions cannot capture interactive effects, i.e. how one agent’s future trajectory might influence or be influenced by others, which limits their usefulness when directly integrated into planning. Among representative approaches, LaPred~\cite{kim2021_lapred} extracts lane–agent features and uses self-supervised attention to identify the most likely lane. Similarly, LAformer~\cite{liu2024laformer} selects relevant lane segments with a dense lane-aware module and refines anchor trajectories to improve temporal consistency. In contrast, PGP~\cite{deo2022_pgp} models the scene as a graph and aggregates context selectively along sampled paths on the graph. Wayformer~\cite{nayakanti2023_wayformer} employs a simpler approach by jointly processing map and agent inputs using a homogeneous attention-based encoder–decoder.

Joint predictors~\cite{jiang2023motiondiffuser, ngiam2022scenetransformer, jfp}, instead model all agents’ futures simultaneously, enabling coordinated and interaction-aware forecasts. For example, MotionDiffuser~\cite{jiang2023motiondiffuser} uses a diffusion model that encodes the scene with a transformer and iteratively de-noises jointly-sampled  trajectories. SceneTransformer~\cite{ngiam2022scenetransformer} embeds agents and road graphs, models interactions via attention, and decodes multiple futures using masked agent–time-step prediction. JFP~\cite{jfp} generates initial per-agent trajectories, computes pairwise potentials for interacting agents, and applies message passing to produce coherent joint predictions.

\subsection{Integrated Prediction and Planning Systems}
Integrating prediction and planning is a key challenge in autonomous driving, as decision-making must account for both ego actions and how others may respond. A comprehensive overview is provided in~\cite{hagedorn2024Integration}, which classifies approaches by the direction of information flow between prediction and planning: \textit{Human Leader}, \textit{Robot Leader}, \textit{Joint}, and \textit{Co-Leader}.

In Human Leader methods, prediction is performed first and planning conditions on it~\cite{dauner2023parting, distelzweig2025perfectprediction, chekroun2023mbappe, Cui2021lookout}. These approaches work well in structured traffic but tend to be overly conservative since the planner reacts to fixed predictions and cannot model mutual adaptation.
Joint approaches optimize prediction and planning within a unified framework~\cite{huang2023gameformer, zheng2025diffusion, pini2022safepathnet, chitta2022transfuser}. While they model multi-agent interaction explicitly, they often assume centralized control or full compliance, which can lead to overconfident behavior.

Our method follows the Co-Leader paradigm, which couples prediction and planning bidirectionally: prediction informs planning, and planning reshapes prediction. This lets the ego vehicle reason about expected cooperation and balance safety and efficiency more effectively, avoiding the over-conservatism of sequential approaches as well as the overconfidence of joint approaches. Few works adopt such reasoning. DTPP~\cite{huang2024dtpp} does so via iterative tree expansion with rule-based ego proposals, an ego-conditioned predictor, and an Inverse Reinforcement Learning-inspired cost. Similarly, \cite{rhinehart2021contingenciesobservations} enables bidirectional reasoning by searching for ego motions that are both likely under a learned multi-agent prediction model as well as satisfy task objectives.

\subsection{Game Theory and Planning}
Game theory provides a mathematical framework for analyzing strategic interactions among rational agents, where each agent’s outcome depends not only on its own choices but also on those of others ~\cite{vonNeumann1944, osborne1994}. A central concept is the Nash equilibrium~\cite{nash1950}, a stable state in which no agent can improve its outcome by unilaterally changing its strategy.

In autonomous driving, game-theoretic reasoning provides a useful framework for modeling interactions. Approaches include non-cooperative games for competitive scenarios such as merging~\cite{sadigh2016}, and Stackelberg games to capture leader–follower dynamics~\cite{li2019gametheoreticmodelingmultivehicleinteractions}. More recently, advanced methods combine game theory with learning-based models: GameFormer~\cite{huang2023gameformer} integrates hierarchical game-theoretic modeling with Transformers for joint interaction prediction and motion planning. Bayesian game formulations explicitly address multi-modal uncertainty in driver intentions~\cite{huang2024integrateddecisionmakingtrajectory}, and mixed-strategy game approaches adapt to diverse social orientations such as defensive or aggressive driving styles~\cite{liu2024mixed}. Embedding such interaction-aware reasoning into planning enables autonomous systems to anticipate, negotiate, and adapt more effectively in dynamic traffic environments.

However, to the best of our knowledge, no approach has unified state-of-the-art predictors and planners within a single game-theoretic framework, leaving a gap for methods that can jointly optimize prediction and planning under interactive, strategic reasoning.
\section{Method}
Our method comprises three components: (\textit{i}) a state-of-the-art marginal prediction model, (\textit{ii}) a sampling-based planner, and (\textit{iii}) a game-theoretic interaction modeling framework with Bayesian confidence estimation. An overview of the approach is provided in Fig.~\ref{fig:biber}. The prediction model produces forecasts of surrounding agents’ intentions by outputting a discrete distribution over future motion trajectories. The planner generates a diverse set of proposals for the ego vehicle, while the game-theoretic framework connects both, allowing the ego to reason about how its actions affect and are affected by other agents.

\noindent Sec.~\ref{sec:marginal_prediction_module} reviews LAformer~\cite{liu2024laformer}, which we adopt as the marginal prediction model. Sec.~\ref{sec:proposal_generator} describes the proposal generator for ego-vehicle trajectories, and Sec.~\ref{sec:biber} details the integration of predictions and ego proposals via the game-theoretic BIBeR framework.

\subsection{Marginal Prediction Module}
\label{sec:marginal_prediction_module}
A key strength of our system is that it works seamlessly with state-of-the-art predictors. In particular, we employ LAformer~\cite{liu2024laformer}, a well-established marginal prediction model.

LAformer takes as input the past trajectories of all agents $X_{1:N_a}^{-t_h:0}$, where $N_a$ denotes the number of surrounding agents, and lane centerlines from the high-definition (HD) map $C_{\text{map}}$. Both are represented in vectorized form and normalized around the target agent. The scene context is encoded and fused in a Global Interaction Graph (GIG) using cross- and self-attention. A temporally dense lane-aware module assigns probabilities to lane segments and selects the top-$k$ candidates, producing lane-aligned motion features for the target agent.  
Conditioned on these features and a latent variable per mode $m$, a multi-modal decoder predicts a set of $M$ candidate future trajectories $\{\hat{Y}^m_{1:T_f}\}_{m=1}^M$ along with their probabilities $\{\pi_m\}_{m=1}^M$, thus forming a discrete distribution over possible futures. Further details can be found in the supplementary material. %(Sec. \ref{subsec:suppl:laformer}).

\subsection{Sampling-based Planner}
\label{sec:proposal_generator}
For proposal generation, we employ SPDM~\cite{distelzweig2025perfectprediction}, which extends the well-known PDM-Closed~\cite{dauner2023parting} planning approach to enable more complex maneuvers such as lane changes. This extension is essential for BIBeR, as it addresses a key limitation of PDM-Closed—the inability to make lane changes. SPDM first samples multiple target positions on the ego vehicle’s current and adjacent lanes, and then fits cubic splines through these points to define lateral motion. Longitudinal motions along these splines are generated using the Intelligent Driver Model (IDM)~\cite{treiber2000congested}  with varying target velocities. Each resulting proposal is subsequently simulated to evaluate its expected outcome.

For surrounding traffic participants, SPDM employs constant-velocity predictions. The generated proposals are then ranked using a cost function that considers factors such as collision, progress, and comfort (details are provided in the supplementary material). %in Sec. \ref{subsec:suppl:spdm}).
This cost function does not account for interaction effects, as it assumes that the ego vehicle cannot influence the behavior of other agents. Instead, BIBeR takes the proposals generated by SPDM and adjusts their ranking based on an iteratively applied explicit interaction model.

\subsection{BIBeR: Bayesian Iterative Best Response}
\label{sec:biber}
Multi-agent planning in autonomous driving can be modeled as a dynamic game, 
where multiple agents interact strategically over a finite horizon. 
Let $\theta_i \in \Theta_i$ denote the strategy of agent $i \in \{1, \dots, N\}$,  with $\Theta_i$ being the feasible strategy set. Each agent seeks to maximize its individual payoff $s(\theta_i)$, which depends both on its own choice and on the strategies of all other agents. A Nash equilibrium is then defined as a strategy profile:
\begin{equation}
    \theta^* = (\theta_1^*, \dots, \theta_{N}^*), \qquad
    \theta_i^* = \arg \max_{\theta_i \in \Theta_i(\theta_{-i}^*)} s(\theta_i) \, ,
\end{equation}
where $\theta_{-i}^*$ denotes the equilibrium strategies of all agents except $i$. At this point, no agent can improve its outcome by unilateral deviation, i.e. only by changing its own strategy.  

\textbf{Iterative Best Response (IBR).} Since solving directly for a Nash equilibrium is computationally intractable in dynamic multi-agent settings, we employ \emph{Iterative Best Response (IBR)}. The core idea is that each agent updates its policy iteratively, always responding optimally to the fixed strategies of others. In the context of trajectory planning, an agent does not generate new candidate trajectories during IBR. Instead, it re-weights a fixed, discrete set of precomputed trajectories based on their expected reward. Our reward function closely aligns with those used in prior work~\cite{dauner2023parting, caesar2022nuplan, Jaeger2025CoRL}.   

\noindent Each agent $i$ produces $M$ candidate trajectories $T_{i,l}$ ($l\in\{1, \dots, M\}$, where $M$ can vary for predicted modes or ego proposals). At iteration $k$, the total reward is computed as
% \begin{align}
% \label{eq:reward}
% R^{k}_{i,l} 
%   &= \sum_{j=0}^{i-1} \left( \frac{1}{M} \sum_{m=0}^{M} 
%       \psi(T_{i,l}, T_{j,m}) \, w^{k+1}_{j,m} \right) \nonumber \\
%   &\quad + \sum_{j=i+1}^{{N}} \left( \frac{1}{M} \sum_{m=0}^{M} 
%       \psi(T_{i,l}, T_{j,m}) \, w^{k}_{j,m} \right) \nonumber \\
%   &\quad + w_p \cdot r_{\text{progress}}(T_{i,l}; \alpha, \beta) \nonumber \\
%   &\quad + w_c \cdot r_{\text{comfort}}(T_{i,l})\quad.
% \end{align}
\begin{equation}
\begin{aligned}
R^{k}_{i,l} &= r_{\text{distance}}(T_{i,l}; w^k, w^{k+1}) \\
&+ w_p \cdot r_{\text{progress}}(T_{i,l}; \alpha, \beta) \\
&+ w_c \cdot r_{\text{comfort}}(T_{i,l}) \quad .
\end{aligned}
\label{eq:reward}
\end{equation}
Here, the distance term consists of pairwise interactions
\begin{equation}
\begin{aligned}
r_{\text{distance}}(T_{i,l}; w^k, w^{k+1}) &= 
\sum_{j=0}^{i-1} \left( \frac{1}{M} \sum_{m=0}^{M} 
      \psi(T_{i,l}, T_{j,m}) \, w^{k+1}_{j,m} \right) \\
&+ \sum_{j=i+1}^{N} \left( \frac{1}{M} \sum_{m=0}^{M} 
      \psi(T_{i,l}, T_{j,m}) \, w^{k}_{j,m} \right) \, ,
\end{aligned}
\label{eq:r_distance}
\end{equation}
where $w_{j, l}^{k}$ weighs trajectory $T_{j,m}$ of agent $j$ at iteration $k$ and $\psi(T_{i,l}, T_{j,m})$ scores the interaction between two trajectories using tunable penalty values $u_c$ and $u_d$ 
\begin{equation}
\label{eq:collision}
\psi(T_{i,l}, T_{j,m}) = 
    \begin{cases}
u_c & \text{if $T_{i,l}$ collides with $T_{j,m}$}, \\
u_d & \text{if $T_{i,l}$ is too close to $T_{j,m}$}, \\
0   & \text{otherwise}\quad .
\end{cases}
\end{equation}
The progress reward $r_{\text{progress}}(T_{i,l}; \alpha, \beta)$ in Eq.~\ref{eq:reward} normalizes the longitudinal and lateral displacements to the interval $[0,1]$ and combines them using the weighting factors $\alpha$ and $\beta$. The comfort term $r_{\text{comfort}}(T_{i,l})$ is binary, taking values in $\{0,1\}$. Each score is scaled by its respective weight, $w_p$ for progress and $w_c$ for comfort.

The trajectory weights of agent $i$ are updated via
\begin{equation}
\label{eq:weight_update}
    w_i^{k+1} = w_i^{k} \, e^{c_i \cdot R_i^{k}} \quad,
\end{equation}
where $c_i$ is a confidence parameter that modulates the influence of the reward on the update. From the updated weights, the trajectory distribution is obtained as
\begin{equation}
\label{eq:dist_update}
    P^{k}(T_{i,l}) = w_{i, l}^{k} \cdot P^{0}(T_{i,l}) \quad,
\end{equation}
where $P^{0}(T_{i,l})$ is the initial distribution over trajectories, i.e. with initial weights.
In practice, IBR is not iterated until full convergence but rather for a fixed number of iterations.

\textbf{Bayesian confidence estimation.} While the multi-agent BIBeR reduces the conservatism of Human Leader approaches (which perform prediction then planning), it may become unsafe if the planner’s assumptions about other agents diverge from reality. In such cases, BIBeR naturally reverts to relying on the initialized marginal predictions—acting more conservatively to maintain safety.

To balance assertiveness and conservatism, BIBeR performs \emph{Bayesian confidence estimation} for each non-ego agent. This measures how well the updated trajectory distribution matches the observed motion compared to the original predictor. To formalize this, we introduce a binary variable $z_t^i \in \{0,1\}$, where $z_t^i = 1$ denotes that BIBeR’s updated trajectory distribution explains the agent’s motion at time $t$, and $z_t^i = 0$ denotes that the original predictor is more accurate. Given the observed state $s_t^i$, the confidence posterior is updated recursively using Bayes’ rule:
\begin{align}
\label{eq:bayes_update}
P(z_t^i \mid s_t^i)=\frac{
P(s_t^i \mid z_{t-1}^i)\, P(z_{t-1}^i \mid s_{t-1}^i)
}{
P(s_t^i \mid z_{t-1}^i)\, P(z_{t-1}^i \mid s_{t-1}^i)
+
P(s_t^i \mid \neg z_{t-1}^i)\, P(\neg z_{t-1}^i \mid s_{t-1}^i)
}\, .
\end{align}
The likelihoods compare the observed motion with the representative trajectories returned by BIBeR and LAformer. Both modules produce trajectory bundles, which we reduce to a single trajectory by selecting the most likely one. The likelihoods are modeled using Gaussian densities
\begin{equation}
P(s_t^i \mid z_{t-1}^i) = \mathcal{N}(s_t^i;\, \hat{s}_{t}^{i,\mathrm{B}},\, \Sigma),
\quad
P(s_t^i \mid \neg z_{t-1}^i) = \mathcal{N}(s_t^i;\, \hat{s}_{t}^{i,\mathrm{Pred}},\, \Sigma)\, ,
\end{equation}
where $\hat{s}_{t}^{i,\mathrm{B}}$ and $\hat{s}_{t}^{i,\mathrm{Pred}}$ denote the BIBeR- and predictor-provided trajectories, respectively, and $\Sigma$ encodes uncertainty.
The final confidence value is then defined as
\begin{equation}
\label{eq:confidence}
c_i = P(z_t^i \mid s_t^i) \quad.
\end{equation}
The resulting confidence enters BIBeR’s weight update (Eq.~\ref{eq:weight_update}). High confidence increases the influence of the
multi-agent planner, while low confidence slows down distribution updates and
drives behavior towards a more conservative prediction-then-planning strategy.
This mutual interaction between planning and confidence estimation enables an
adaptive trade-off between assertiveness and safety. We provide pseudo-code of the entire approach in Alg. \ref{alg:biber}. Further details are in the supplementary material. % (Sec.~\ref{sec:suppl:ibr-details}).

\begin{algorithm}[h]
\caption{Bayesian Iterative Best Response (BIBeR)}\label{alg:biber}
\begin{algorithmic}[1]
\Require Context $C = \{X_{1:N_a}^{-t_h:0}, C_{\text{map}}\}$
\Ensure Ego trajectory $Y^{t}_{\text{ego}}$
\State $P^{0}_{\text{agents}} \gets \text{LAformer}(C)$ \Comment{Prediction, Sec.~\ref{sec:marginal_prediction_module}}
\State $P^{0}_{\text{ego}} \gets \text{Proposal}(C)$ \Comment{Proposal generation, Sec.~\ref{sec:proposal_generator}}
\\
\State $c \gets \text{BayesianConfidence}(C)$ \Comment{Eq.~\ref{eq:bayes_update},~\ref{eq:confidence}}
\vspace{0.5em}

%\State $r_\text{{distance}} \gets \text{Distances}(P^{0}_{\text{agents}}, P^{0}_{\text{ego}})$
\State $r_\text{{distance}} \gets \text{Distances}(\cdot)$ 
\Comment{Eq.~\ref{eq:r_distance}}
%\State $r_\text{{distance}} \gets \text{Distances}(\{{T}_{i,l}, {T}_{j,m}\}^{M}_{i=l=j=m=0})$ 
%\State $r_\text{{progress}} \gets \text{Progress}(P^{0}_{\text{ego}}, C)$
%\State $r_{\text{comfort}} \gets \text{Comfort}(P^{0}_{\text{ego}})$
\State $r_\text{{progress}} \gets \text{Progress}(\cdot)$
\State $r_{\text{comfort}} \gets \text{Comfort}(\cdot)$
\vspace{0.5em}

\While{not converged}
    \For{each agent $i \in N$}
        \For{each trajectory $T_{i,m}$}
            \State $R^{k}_{i,m} \gets %\text{TotalReward}(\cdot)$ \Comment{Eq.~\ref{eq:reward}}
            \text{Reward}(r_\text{dist}, r_\text{prog}, r_\text{comf})$ \Comment{Eq.~\ref{eq:reward}}
            \State $w_{i,m}^{k+1} \gets w_{i,m}^{k} \cdot e^{\, c_i \cdot R_{i,m}^{k}}$ \Comment{Eq.~\ref{eq:weight_update}}
        \EndFor
    \EndFor
\EndWhile
\vspace{0.5em}

\State $P^{k} \gets \text{UpdateDistributions}(P^{0}_{\text{agents}}, P^{0}_{\text{ego}}, w^{k})$ \Comment{Eq.~\ref{eq:dist_update}}
\State \Return $Y^{t}_{\text{ego}} \gets \text{BestTrajectory}(P^{k}_{\text{ego}})$
\end{algorithmic}
\end{algorithm}
\vspace{-1em}
\section{Evaluation}
\subsection{Evaluation Benchmarks}
\label{sec:benchmarks}
We build on the nuPlan framework~\cite{caesar2022nuplan}, a closed-loop simulator based on real-world driving data. For evaluation, we use the Val14 benchmark~\cite{dauner2023parting}, which comprises 1,118 scenarios from the nuPlan validation set, covering up to 100 examples per type across 14 scenario categories.
We further consider the Test14 benchmark~\cite{jcheng2023plantf}, which is divided into Test14-random and Test14-hard. Test14-random consists of randomly sampled scenarios from each category, whereas Test14-hard is constructed by running 100 scenarios of each type with the state-of-the-art rule-based planner PDM-Closed~\cite{dauner2023parting} and selecting the 20 most challenging cases per type based on performance.

In addition, we include the interPlan benchmark~\cite{hallgarten2024interplan}, an out-of-distribution set of 80 manually created complex scenarios such as maneuvering around parked vehicles, reacting to jaywalkers, and executing lane changes under dense traffic. For analysis, we consider both the full interPlan set (80 scenarios) and a dedicated lane-change subset. The lane-change subset contains 30 scenarios with particularly rich ego vehicle and traffic interaction, covering low, medium, and high traffic densities with 10 cases for each.

In most benchmark settings, surrounding agents are either non-reactive (NR), following pre-recorded trajectories and ignoring deviations of the ego agent from the expert trajectory, or reactive (R), using the Intelligent Driver Model (IDM)~\cite{treiber2000congested} to respond to the ego agent’s behavior. However, IDM agents do not sufficiently capture human driving behavior~\cite{hagedorn2025plannersmeetreality}—a limitation that is particularly important for our approach, which explicitly models interactions and relies on realistic reactive responses. To address this, we follow~\cite{hagedorn2025plannersmeetreality} and additionally evaluate using SMART-reactive (SR) agents. The SMART model~\cite{wu2024smart} generates behaviors that are both reactive and more human-like than IDM or log-replay methods, providing a more realistic evaluation environment. For more details, we refer to the supplementary material. % (Sec. \ref{sec:suppl:benchmarks}).

\subsection{Implementation Details}
\label{sec:implementation}
We use $M_{\text{a}} = 5$ predictions for each surrounding agent from the prediction model. We provide two versions of BIBeR: 
\begin{itemize}
    \item \textit{BIBeR}: operates on the first $M_{\text{e}} = 128$ ego proposals generated by SPDM, without any filtering.
    \item \textit{BIBeR-CV}: ego proposals generated by SPDM are pre-filtered using constant-velocity (CV) predictions of surrounding agents, the motivation for this version of BIBeR is described in Sec.~\ref{sec:cv_prefiltering}.
\end{itemize}
Both the proposals and predictions cover a future time horizon of $t_f = 4\,\text{s}$.
The prediction model predicts eight future positions at $2\,\text{Hz}$, which are upsampled to $10\,\text{Hz}$, resulting in a total of 40 positions. Ego proposals are generated directly at $10\,\text{Hz}$. The ego agent is always updated first, and we perform $10$ iterations of IBR. The runtime of BIBeR is largely determined by its prediction and planning modules. Detailed runtime analysis is provided in the supplementary material. % (Sec. \ref{subsec:suppl:runtime}).

\section{Results}
\begin{table*}[tp]
    \centering
    \resizebox{1\textwidth}{!}{
    \begin{tabular}{llrrrrrrrr}
        \toprule
        \multicolumn{2}{c}{Planner}
        & \multicolumn{2}{c}{Val14}
        & \multicolumn{2}{c}{Test14-hard}
        & \multicolumn{2}{c}{Test14-random}
        & \multicolumn{1}{c}{interPlan}
        & \multicolumn{1}{c}{interPlanLC} \\
        \cmidrule(lr){1-2}\cmidrule(lr){3-4}\cmidrule(lr){5-6}\cmidrule(lr){7-8}\cmidrule(lr){9-9}\cmidrule(lr){10-10}
        Integration & Method
        & CLS-NR $\uparrow$ & CLS-R $\uparrow$
        & CLS-NR $\uparrow$ & CLS-R $\uparrow$
        & CLS-NR $\uparrow$ & CLS-R $\uparrow$
        & CLS-R $\uparrow$
        & CLS-R $\uparrow$ \\
        \midrule

        % NONE
        \multirow{5}{*}{None} 

        & Urban Driver~\cite{scheel2022urban} 
        & 68.57 & 64.11 
        & 50.40 & 49.95 
        & 51.83 & 67.15
        & 4.00 & 9.65 \\

        & GC-PGP~\cite{hallgarten2023prediction} 
        & 58.40 & 63.82
        & 43.22 & 39.36 
        & 55.99 & 51.39 
        & 14.55 & 34.99 \\

        & PlanCNN~\cite{renz2022plant} 
        & 73.00 & 72.00 
        & 49.40 & 52.20
        & 69.70 & 67.50 
        & --- & 58.35 \\

        & PlanTF~\cite{jcheng2023plantf} 
        & 84.27 & 76.95 
        & 69.70 & 61.61 
        & 85.62 & 79.58
        & 30.53 & 55.18 \\

        & PDM-Open~\cite{dauner2023parting} 
        & 53.53 & 54.24 
        & 33.51 & 35.83 
        & 52.81 & 57.23
        & --- & 23.35 \\

        \midrule
        % HUMAN LEADER
        \multirow{2}{*}{Human Leader} 
        & PDM-Closed~\cite{dauner2023parting}  
        & \textbf{92.84} & \underline{92.12}
        & 65.08 & 75.19 
        & 90.05 & \underline{91.63}
        & 41.88 & 61.55 \\

        & PDM-Hybrid~\cite{dauner2023parting}  
        & 92.77 & 92.11 
        & 65.99 & 76.07 
        & 90.10 & 91.28
        & 41.61 & 61.72 \\

        \midrule
        % JOINT
        \multirow{2}{*}{Joint} 
        & GameFormer~\cite{huang2023gameformer}  
        & 79.94 & 79.78 
        & 68.70 & 67.05 
        & 83.88 & 82.05
        & 21.26 & 30.49 \\

        & DiffusionPlanner~\cite{zheng2025diffusion}  
        & 89.87 & 82.80 
        & 75.99 & 69.22 
        & 89.19 & 82.93
        & 25.76 & 25.96 \\

        \midrule
        % CO-LEADER
        \multirow{3}{*}{Co-Leader} 
        & DTPP~\cite{huang2024dtpp}  
        & 71.66 & 66.09 
        & 60.11 & 63.66 
        & --- & 74.94
        & 30.32 & 67.88 \\

        & \textbf{BIBeR} (Ours) 
        & 90.07 & 89.79
        & \underline{73.00} & \underline{79.57}
        & \underline{91.12} & 90.92
        & \underline{55.59} & \underline{81.17} \\

        & \textbf{BIBeR-CV} (Ours) 
        & \underline{91.92} & \textbf{92.75} 
        & \textbf{80.30} & \textbf{83.00}
        & \textbf{93.18} & \textbf{93.01}
        & \textbf{60.05} & \textbf{84.53} \\
        \bottomrule
    \end{tabular}
    }
    \caption{Comparison of planning performance on the nuPlan benchmarks across different approaches. We report closed-loop non-reactive (CLS-NR) and closed-loop reactive (CLS-R) scores for Val14, Test14-hard, and Test14-random splits, as well as reactive scores on the interPlan and interPlanLC scenarios. Higher values indicate better performance. The best score is shown in \textbf{bold}, and the second-best score is \underline{underlined}.}
    \vspace{-0.8cm}
    \label{tab:performance}
\end{table*}
Our evaluation is structured in two stages: (\textit{i}) experiments conducted under the standard nuPlan simulation settings (NR and R), and (\textit{ii}) experiments using learned traffic agents (SR), detailed in Sec. \ref{sec:benchmarks}.

\subsection{Constant-Velocity Proposal Prefiltering}
\label{sec:cv_prefiltering}
\begin{figure}[t]
    \center
    \includegraphics[width=0.9\columnwidth]{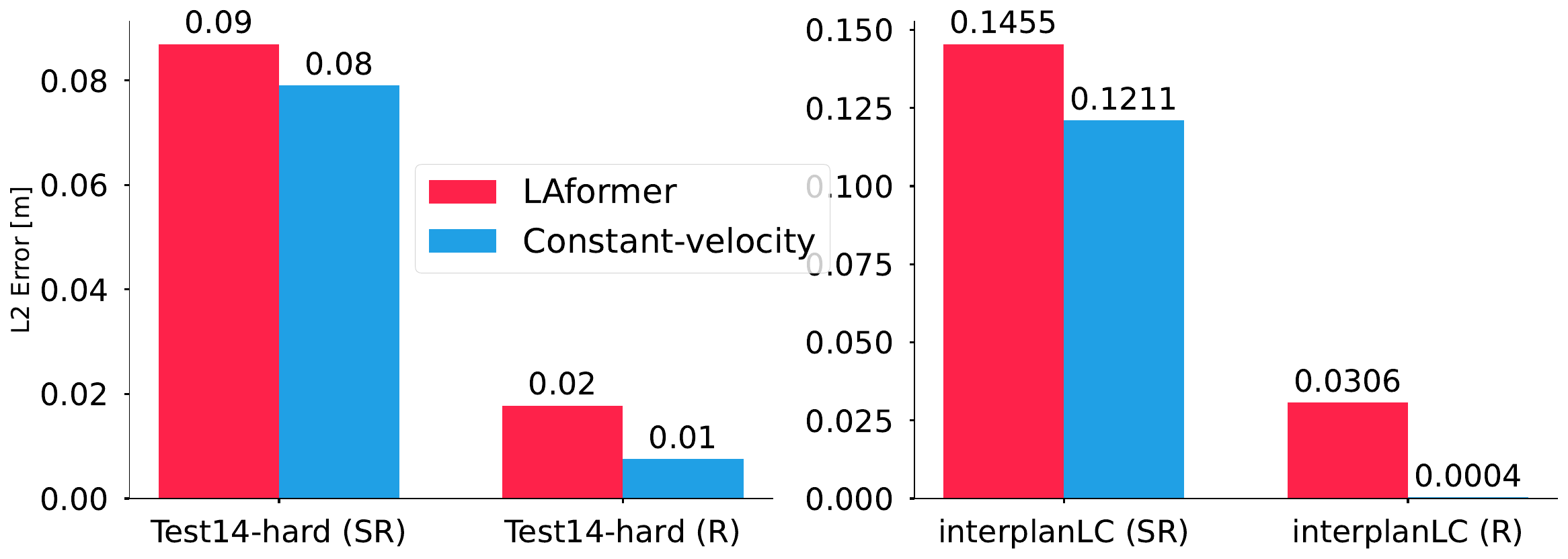}
    \caption{Prediction error at a $0.1$s horizon for LAformer and constant-velocity under reactive (R) and SMART-reactive (SR) settings.\textbf{ Left:} Errors on the Test14-hard benchmark. \textbf{Right:} Errors on the interPlanLC benchmark.}
    \label{fig:pred_error}
    \vspace{-0.6cm}
\end{figure}

The core contribution of BIBeR is a principled integration of learning-based prediction into planning. In the standard nuPlan reactive setting, surrounding agents are simulated using IDM, which does not realistically reflect human driving. Thus, learned prediction models cannot reliably predict IDM trajectories, while constant-velocity (CV) predictions are much closer to IDM behavior but are unimodal and inadequately responsive to the ego vehicle, making them unsuitable as direct predictions for BIBeR. To illustrate this, Fig.~\ref{fig:pred_error} shows the prediction error at a horizon of 0.1 seconds, the most critical in nuPlan due to the 10Hz replanning frequency. Unsurprisingly, CV predictions outperform LAformer across all simulation settings for this horizon, but the relative error decreases in the learned traffic agent (SR) setting compared to the reactive IDM (R) setting.
\noindent This motivates the design of our evaluation: in reactive IDM and non-reactive settings, we evaluate BIBeR-CV using CV predictions as proposal filters to partially compensate for the unrealistic IDM behavior while still allowing the ego vehicle to influence other agents. In the learned traffic agent setting, we evaluate BIBeR without filtering as learned traffic agents provide a much closer approximation of human driving behavior and enable proper assessment of integrated learning-based prediction.
\subsection{Evaluation with IDM Traffic Agents}
We report the results of BIBeR-CV in standard nuPlan settings in Tab. \ref{tab:performance}. BIBeR-CV consistently outperforms all other approaches across most benchmarks with the exception of Val14 in the NR setting, where it is slightly worse than PDM-Closed.
The performance improvements are particularly pronounced on benchmarks with more complex scenarios such as interPlan, interPlanLC, and Test14-hard. Notably, BIBeR-CV achieves a $24\%$ boost over DTPP~\cite{huang2024dtpp}, another Co-Leader approach, on interPlanLC, highlighting its ability to handle interactive and challenging scenarios more effectively than baselines. BIBeR without filtering, i.e. relying solely on learned predictions, also outperforms most other approaches on majority of benchmarks, showing a strong advantage in interactive scenarios.

\subsection{Evaluation with Learned Traffic Agents}
Planners are often evaluated in simulators where the behavior of surrounding agents is unrealistic, which can lead to biased or misleading performance assessments. This issue is particularly severe for BIBeR, since it relies on a learning-based predictor. To mitigate this issue, we additionally evaluate BIBeR in an environment with learned traffic agents (see Sec. \ref{sec:benchmarks} for details), with results presented in Tab. \ref{tab:performance-sr}.
The results closely follow the trends observed in Tab.~\ref{tab:performance}: BIBeR exhibits particularly strong performance on benchmarks involving complex and highly interactive scenarios, such as Test14-hard and interPlanLC. While the improvement on Test14-hard is modest, the advantage becomes substantially more pronounced on the strongly interactive interPlanLC benchmark, where BIBeR achieves nearly a $30\%$ performance improvement over PDM-Hybrid, the second-best baseline. We provide further results in the supplementary.

\subsection{Importance of BIBeR} 
% Preamble (ensure these are included)
% \usepackage{booktabs}
% \usepackage{multirow}
% \usepackage{caption}

\begin{figure*}[t]
\captionsetup{skip=0pt}
\centering

% ===================== Table 2 (LEFT, spans both rows) =====================
\begin{minipage}[t]{0.48\textwidth}
\vspace{0pt}
\centering
\resizebox{\textwidth}{!}{%
\begin{tabular}{llrrr}
    \toprule
    \multicolumn{2}{c}{Planner}
    & \multicolumn{3}{c}{CLS-SR $\uparrow$} \\
    \cmidrule(lr){1-2}\cmidrule(lr){3-5}
    Integration & Method
    & Val14 & Test14-hard
    & interPlanLC \\
    \midrule

    \multirow{5}{*}{None}
    & Urban Driver~\cite{scheel2022urban}
    & 43.45 & 38.38 & 4.87 \\
    & GC-PGP~\cite{hallgarten2023prediction}
    & 50.82 & 40.96 & 14.95 \\
    & PlanCNN~\cite{renz2022plant}
    & 64.70 & 50.78 & 22.78 \\
    & PlanTF~\cite{jcheng2023plantf}
    & 71.78 & 57.50 & 37.67 \\
    & PDM-Open~\cite{dauner2023parting}
    & 53.49 & 38.10 & 15.59 \\

    \midrule
    \multirow{2}{*}{Human Leader}
    & PDM-Closed~\cite{dauner2023parting}
    & \underline{89.14} & \underline{73.99} & 54.06 \\
    & PDM-Hybrid~\cite{dauner2023parting}
    & \textbf{89.44} & 72.15 & \underline{54.28} \\

    \midrule
    \multirow{2}{*}{Joint}
    & GameFormer~\cite{huang2023gameformer}
    & 78.05 & 62.48 & 37.98 \\
    & DiffusionPlanner~\cite{zheng2025diffusion}
    & 77.80 & 63.09 & 29.78 \\

    \midrule
    \multirow{2}{*}{Co-Leader}
    & DTPP~\cite{huang2024dtpp}
    & 62.41 & 54.65 & 45.94 \\
    & \textbf{BIBeR} (Ours)
    & 87.36 & \textbf{75.72}
    & \textbf{70.03} \\

    \bottomrule
\end{tabular}%
}
\captionof{table}{Performance comparison on the nuPlan benchmarks in the SMART Reactive (CLS-SR) setting. Results are reported for Val14, Test14-hard, and interPlanLC, with higher scores indicating better closed-loop performance.}
\label{tab:performance-sr}
\end{minipage}
\hfill
% ===================== RIGHT COLUMN: Table 3 (top) + Table 4 (bottom) =====================
\begin{minipage}[t]{0.48\textwidth}
\vspace{0pt}

% --------------------- Table 3 (TOP RIGHT) ---------------------
\begin{minipage}[t]{\textwidth}
\vspace{0pt}
\centering
\resizebox{\textwidth}{!}{%
\begin{tabular}{lcccc}
    \toprule
    \multirow{2}{*}{Method}
    & \multicolumn{4}{c}{CLS-R $\uparrow$} \\
    \cmidrule(lr){2-5}
    & Val14 & Test14-hard & Test14-random & interPlanLC \\
    \midrule
    SPDM~\cite{distelzweig2025perfectprediction}
    & 92.72 & 80.75 & 92.54 & 84.07 \\
    \textbf{BIBeR-CV} (Ours)
    & \textbf{92.75} & \textbf{83.00} & \textbf{93.01} & \textbf{84.53} \\
    \bottomrule
\end{tabular}%
}
\captionof{table}{Comparison between SPDM and BIBeR-CV on the nuPlan and interPlan benchmarks.}
\label{tab:spdm_comparison_nr_r}
\end{minipage}

% --------------------- Table 4 (BOTTOM RIGHT) ---------------------
\begin{minipage}[t]{\textwidth}
\vspace{0pt}
\centering
\resizebox{\textwidth}{!}{%
\begin{tabular}{lcccc}
    \toprule
    \multirow{2}{*}{Method}
    & \multicolumn{4}{c}{CLS-SR $\uparrow$} \\
    \cmidrule(lr){2-5}
    & Val14 & Test14-hard & Test14-random & interPlanLC \\
    \midrule
    SPDM~\cite{distelzweig2025perfectprediction}
        & \textbf{88.53} & 75.01 & \textbf{89.50} & 63.40 \\
    \textbf{BIBeR} (Ours)
        & 87.36 & \textbf{75.72} & 86.50 & \textbf{70.03} \\
    \bottomrule
\end{tabular}%
}
\captionof{table}{Comparison between SPDM and BIBeR-CV on nuPlan and interPlan benchmarks in SMART Reactive (CLS-SR) simulation.}
\label{tab:spdm_comparison_sr}
\end{minipage}

\end{minipage}

\vspace{-0.6cm}
\end{figure*}
In BIBeR, we employ the proposal set generated by SPDM~\cite{distelzweig2025perfectprediction}. SPDM is not only a proposal generator but—following PDM~\cite{dauner2023parting}—also selects the best proposal from this set. Both SPDM and PDM rely on constant-velocity predictions for surrounding agents and therefore neglect interactions between the ego vehicle and other traffic participants.
We directly compare against SPDM to evaluate whether and to what extent BIBeR provides an advantage over this interaction-agnostic approach.

We analyze both versions of BIBeR. The results for SPDM and BIBeR-CV are presented in Tab.~\ref{tab:spdm_comparison_nr_r}. We evaluate BIBeR-CV in non-reactive and reactive settings. The results demonstrate that BIBeR provides an improvement over SPDM. Performance gains are modest on benchmarks with routine driving scenarios such as Val14 and Test14-random. However, in more challenging scenarios, such as Test14-hard, BIBeR shows a substantial improvement over SPDM. On interPlanLC, the improvement is less pronounced since SPDM already performs very strong in this setting.
A similar trend is observed for BIBeR without prefiltering under learned traffic agents, with results shown in Tab.~\ref{tab:spdm_comparison_sr}. Improvements are noticeable in the more challenging and interactive scenarios, Test14-hard and interPlanLC. While the gain on Test14-hard is moderate, BIBeR demonstrates a substantial improvement on the interPlan lane-change scenarios, achieving an $11\%$ increase over SPDM. These results highlight that the BIBeR framework significantly contributes to better performance in complex and interactive driving scenarios. We provide a more detailed analysis in the supplementary. % (Sec. \ref{subsec:suppl:comparison-spdm} and Sec.~\ref{subsec:suppl:additional-res-biber}).

\subsection{Ablation Studies}
\begin{figure}[t]
\centering
\captionsetup{skip=0pt}

% ================= LEFT: TABLE =================
\begin{minipage}[t]{0.48\columnwidth}
\vspace{0pt}
\centering

\resizebox{\textwidth}{!}{%
\begin{tabular}{lcccc}
    \toprule
    \multirow{2}{*}{Dynamics Model} 
    & \multicolumn{4}{c}{CLS-R $\uparrow$} \\
    \cmidrule(lr){2-5}
    & Val14
    & Test14-hard
    & Test14-random
    & interPlan \\
    \midrule

    \multirow{2}{*}{\shortstack[l]{Marginal Predictor\\(LAformer~\cite{liu2024laformer})}}
    & \multirow{2}{*}{\textbf{89.79}}
    & \multirow{2}{*}{\textbf{79.57}}
    & \multirow{2}{*}{\textbf{90.92}}
    & \multirow{2}{*}{\textbf{55.59}} \\
    & & & & \\
    \midrule

    Joint Predictor~\cite{huang2023gameformer}
    & 81.85 & 70.92 & 82.27 & 49.37 \\

    \bottomrule
\end{tabular}%
}

\captionof{table}{Closed-loop reactive (CLS-R) performance of BIBeR when paired with different prediction models. Results are reported on Val14, Test14-hard, Test14-random, and interPlan, demonstrating the impact of marginal vs. joint multi-agent prediction on downstream planning quality.}
\label{tab:predictor}

\end{minipage}
\hfill
% ================= RIGHT: FIGURE =================
\begin{minipage}[t]{0.48\columnwidth}
\vspace{0pt}
\centering

\includegraphics[width=\textwidth]{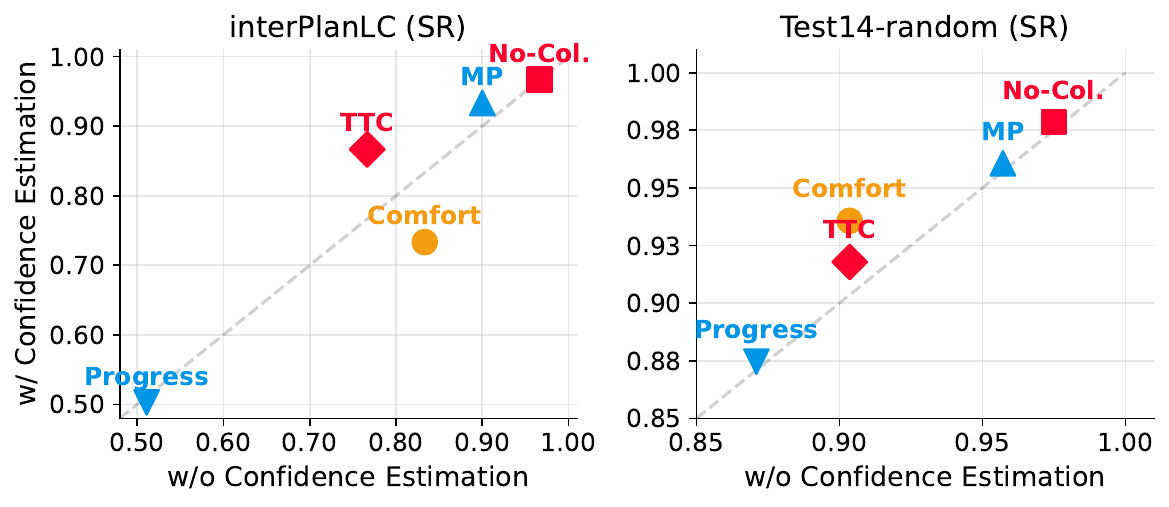}

\captionof{figure}{Comparison of nuPlan subscores with confidence estimation enabled (y-axis) vs. disabled (x-axis). \textbf{Left:} interPlanLC SR. \textbf{Right:} Test14-random SR. BIBeR with LAformer is used.}
\label{fig:confidence}

\end{minipage}

\vspace{-0.6cm}
\end{figure}
% \begin{figure}[t]
% \captionsetup{skip=3pt}
%     \center
%     \includegraphics[width=1\columnwidth]{figures/confidence_comparison.pdf}
%     \caption{Comparison of nuPlan subscores with confidence estimation enabled (y-axis) versus disabled (x-axis). \textbf{Left:} evaluation on interPlanLC SMART reactive (SR) benchmark. \textbf{Right:} evaluation on Test14-random SMART reactive (SR) benchmark. The evaluation uses BiBeR with LAformer as the prediction model. Deviations from the diagonal indicate changes in performance. Other evaluation metrics are omitted, as they show no change in performance and are not relevant to this analysis.}
%     \vspace{-0.5cm}
%     \label{fig:confidence}
% \end{figure}
\textbf{Comparison of prediction models.} We consider two distinct approaches to modeling scene dynamics. The first, LAformer ~\cite{liu2024laformer}, is a marginal predictor, i.e., it predicts the future trajectory of each agent independently of others. The second is a joint prediction model~\cite{huang2023gameformer}, which forecasts the trajectories of all agents simultaneously. The results of BIBeR under both models are reported in Tab.~\ref{tab:predictor}. The joint predictor performs worse than the marginal predictor across all benchmarks. The joint predictor already commits to a specific interaction mode for all agents. This forces BIBeR to follow that mode, even if it is not the one that best serves the ego’s objective. In contrast, marginal prediction preserves indifference over modes, allowing BIBeR to select the interaction mode that is most appropriate for the ego during planning.

\textbf{Influence of Bayesian confidence estimation.} The intuition behind Bayesian confidence estimation is that if past updated predictions have frequently been incorrect, the model is likely to remain unreliable in the near future. Consequently, it should behave more conservatively, applying updates more gradually to ensure a higher level of safety in subsequent decisions.

We analyze the impact of Bayesian confidence estimation, with the results presented in Fig.~\ref{fig:confidence}. In this experiment, we compare our approach with and without confidence estimation. Evaluations are conducted on the interPlan lane change scenarios using learned traffic agents, where this mechanism is expected to demonstrate its full potential, as these scenarios involve highly interactive and uncertain behaviors that require accurate estimation of confidence. Additionally, we evaluate on the Test14-random benchmark, which contains a diverse set of average driving situations with lower interaction complexity.

As shown in Fig.~\ref{fig:confidence}, incorporating confidence estimation improves the time-to-collision (TTC) score on both benchmarks, with a significantly stronger improvement observed on the interPlan lane change benchmark. This can be attributed to the higher degree of interaction and uncertainty during lane changes, where small prediction errors can quickly lead to unsafe maneuvers. In such situations, the confidence estimation mechanism effectively moderates overly-aggressive updates, resulting in safer and more stable planning behavior. No consistent trend is observed for the other subscores. Overall, these results indicate that Bayesian confidence estimation plays a crucial role in improving safety, particularly in highly interactive and safety-critical scenarios.

\begin{table}[t]
\centering
\resizebox{1.0\columnwidth}{!}{
\begin{tabular}{cccccccccc}
\toprule
\multirow{2}{*}{Iteration } 
& \multirow{2}{*}{Score $\uparrow$} 
& \multicolumn{7}{c}{Subscores $\uparrow$}
& \multirow{2}{*}{Runtime (ms) $\downarrow$} \\
\cmidrule(lr){3-9} 
& & No-Col. & TTC & Progress & Comfort & Drivable & Direction & Speed & \\
\midrule
0 & 61.88 & 99.29 & 97.50 & 46.97 & 65.71 & 99.64 & 100.00 & 99.84 & 172\\
1 & 89.89 & 98.04 & 94.64 & 86.69 & 90.00 & 99.64 & 100.00 & 99.74 & 183\\
2 & 90.43 & 98.57 & 95.00 & 86.97 & 90.00 & 99.64 & 100.00 & 99.74 & 189\\
3 & 90.52 & 98.93 & 95.00 & 86.91 & 91.79 & 99.64 & 99.82 & 99.74 & 198\\
$\ldots$ & $\ldots$ & $\ldots$ & $\ldots$ & $\ldots$ & $\ldots$ & $\ldots$ & $\ldots$ & $\ldots$ & $\ldots$\\
10 & \textbf{90.92} & 98.93 & 95.00 & 87.10 & 92.14 & 99.64 & 100.00 & 99.74 & 257\\
\midrule
\end{tabular}}
\caption{Overall score, individual subscores, and per-scenario runtime across varying numbers of iterations in Iterative Best Response (IBR) on the Test14-random reactive benchmark. Results are obtained using BIBeR with LAformer as the prediction model. }
\vspace{-0.8cm}
\label{tab:iterations}
\end{table}
\textbf{Influence of iterations.} We present the results of Iterative Best Response (IBR) with varying numbers of iterations in Tab.~\ref{tab:iterations}, reporting performance improvements after each update of the trajectory distributions. The results show that the most significant improvement occurs after the first iteration.
It can be observed that after the first iteration, progress score increase substantially, while the collisions (No-Col.) and time-to-collision (TTC) scores decrease slightly. This behavior arises because, before the first iteration, all ego proposals are uniformly distributed, and the initially selected proposal corresponds to a very conservative, i.e. low-progress but safe trajectory. Another reason is that the exponential update leads to a large initial jump. Starting from the first iteration, as the proposal distribution is updated, the ego vehicle begins to select trajectories that balance safety and progress more effectively, resulting in higher progress and slightly increased collision and TTC metrics.
No noticeable changes are observed in drivable-area-compliance, driving-direction-compliance, and speed-limit-compliance, as all proposals are generated to inherently satisfy these constraints.

\section{Conclusion}
In this work, we introduced BIBeR, a game-theoretic approach that enables the ego vehicle to both react to and influence the behavior of other agents. In addition, the proposed Bayesian confidence estimation enables the ego to regulate own driving behavior. Our method specifically addresses interactive driving scenarios, where anticipating the intentions and reactions of others is crucial. BIBeR significantly outperforms existing approaches in such scenarios, demonstrating that the mutual integration of prediction and planning can improve driving performance compared to sequential approaches. Nonetheless, evaluating integrated, learning-based prediction-planners remains challenging: interactive benchmarks like interPlan are limited, and simulated traffic agents capture only a subset of possible behaviors. Expanding the evaluation to more diverse and highly interactive scenarios is essential to fully assess the potential of integrated approaches.

\bibliographystyle{splncs04}
\bibliography{main}

\newcommand{\maketitlesupplementary}{\begin{center}\LARGE Supplementary Material\end{center}}
\clearpage
\setcounter{page}{1}

\maketitlesupplementary

% \section{Rationale}
% \label{sec:rationale}
% % 
% Having the supplementary compiled together with the main paper means that:
% % 
% \begin{itemize}
% \item The supplementary can back-reference sections of the main paper, for example, we can refer to \cref{sec:intro};
% \item The main paper can forward reference sub-sections within the supplementary explicitly (e.g. referring to a particular experiment); 
% \item When submitted to arXiv, the supplementary will already included at the end of the paper.
% \end{itemize}
% % 
% To split the supplementary pages from the main paper, you can use \href{https://support.apple.com/en-ca/guide/preview/prvw11793/mac#:~:text=Delete%20a%20page%20from%20a,or%20choose%20Edit%20%3E%20Delete).}{Preview (on macOS)}, \href{https://www.adobe.com/acrobat/how-to/delete-pages-from-pdf.html#:~:text=Choose%20%E2%80%9CTools%E2%80%9D%20%3E%20%E2%80%9COrganize,or%20pages%20from%20the%20file.}{Adobe Acrobat} (on all OSs), as well as \href{https://superuser.com/questions/517986/is-it-possible-to-delete-some-pages-of-a-pdf-document}{command line tools}.

\section{BIBeR}

\subsection{Iterative Best Response}
\label{sec:suppl:ibr-details}
\begin{figure}[h]
    \centering
    \begin{subfigure}[b]{0.19\columnwidth}
        \centering
        \includegraphics[width=\textwidth]{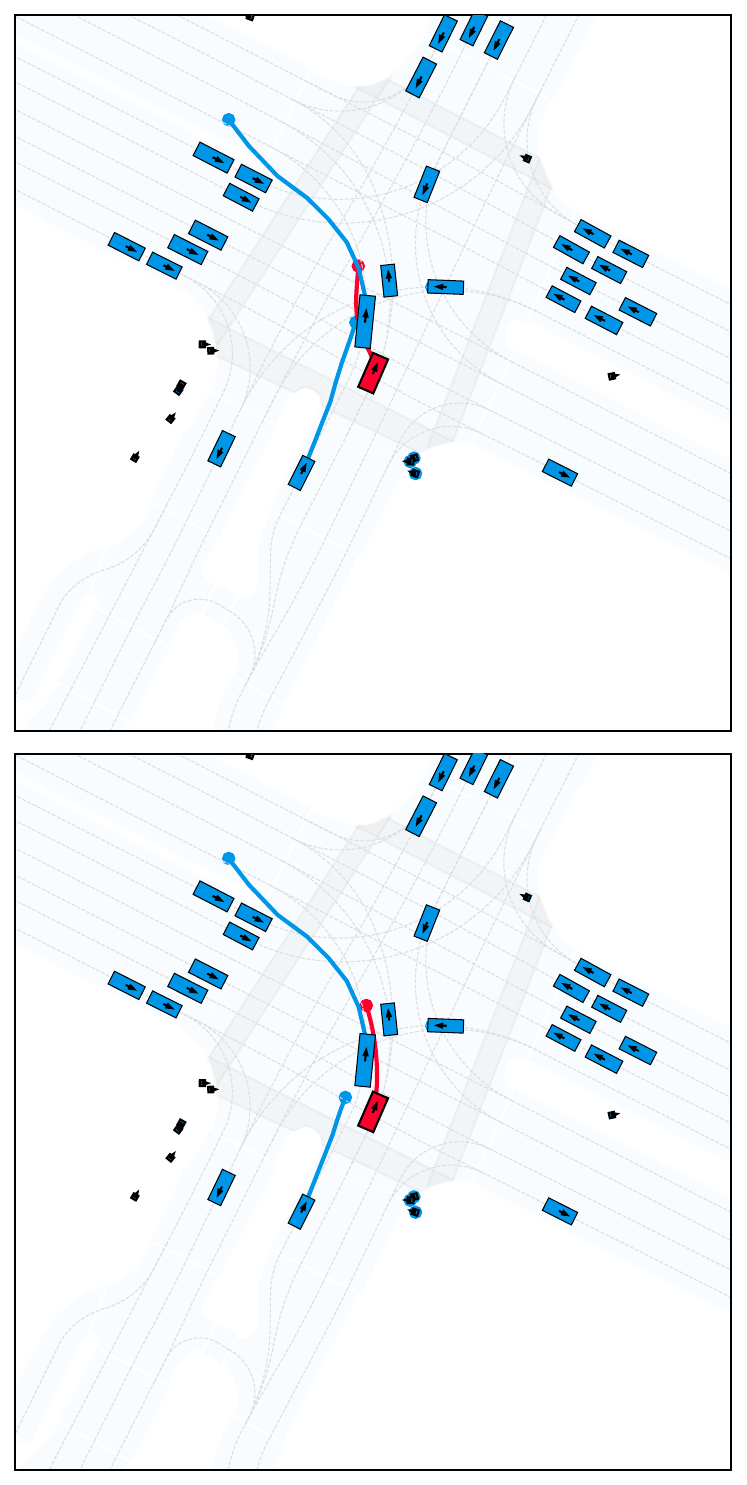}
        \caption{Scenario 1}
    \end{subfigure}
    \begin{subfigure}[b]{0.19\columnwidth}
        \centering
        \includegraphics[width=\textwidth]{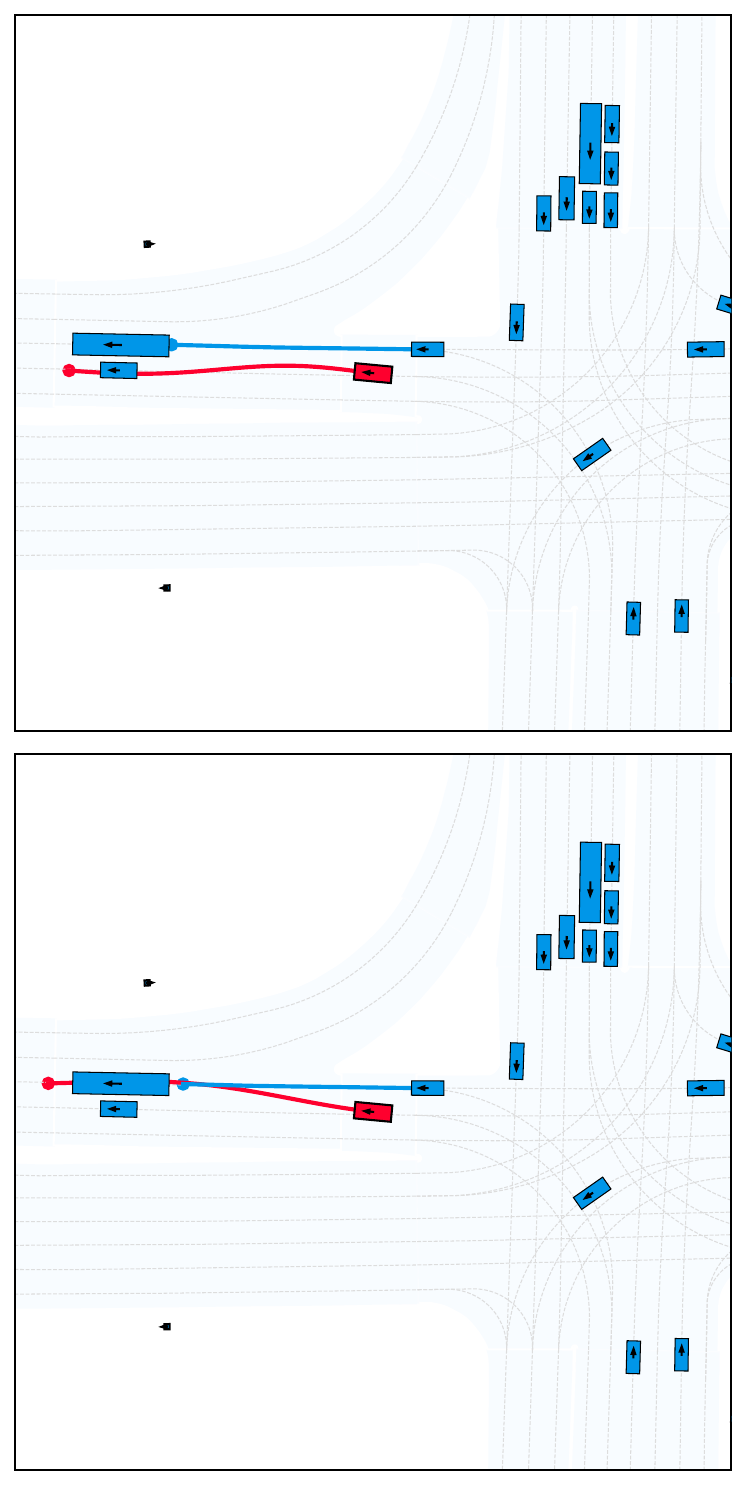}
        \caption{Scenario 2}
    \end{subfigure}
        \begin{subfigure}[b]{0.19\columnwidth}
        \centering
        \includegraphics[width=\textwidth]{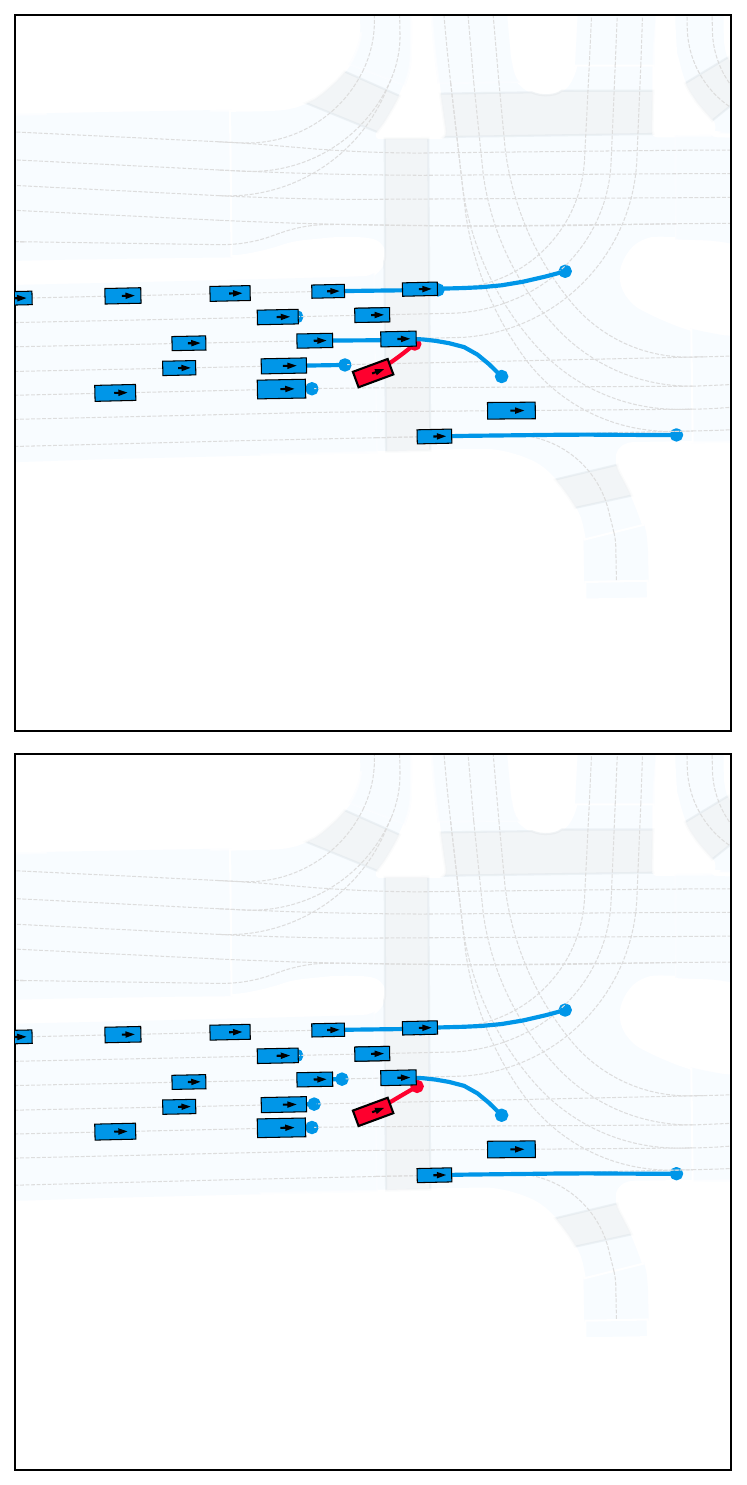}
        \caption{Scenario 3}
    \end{subfigure}
    \begin{subfigure}[b]{0.19\columnwidth}
        \centering
        \includegraphics[width=\textwidth]{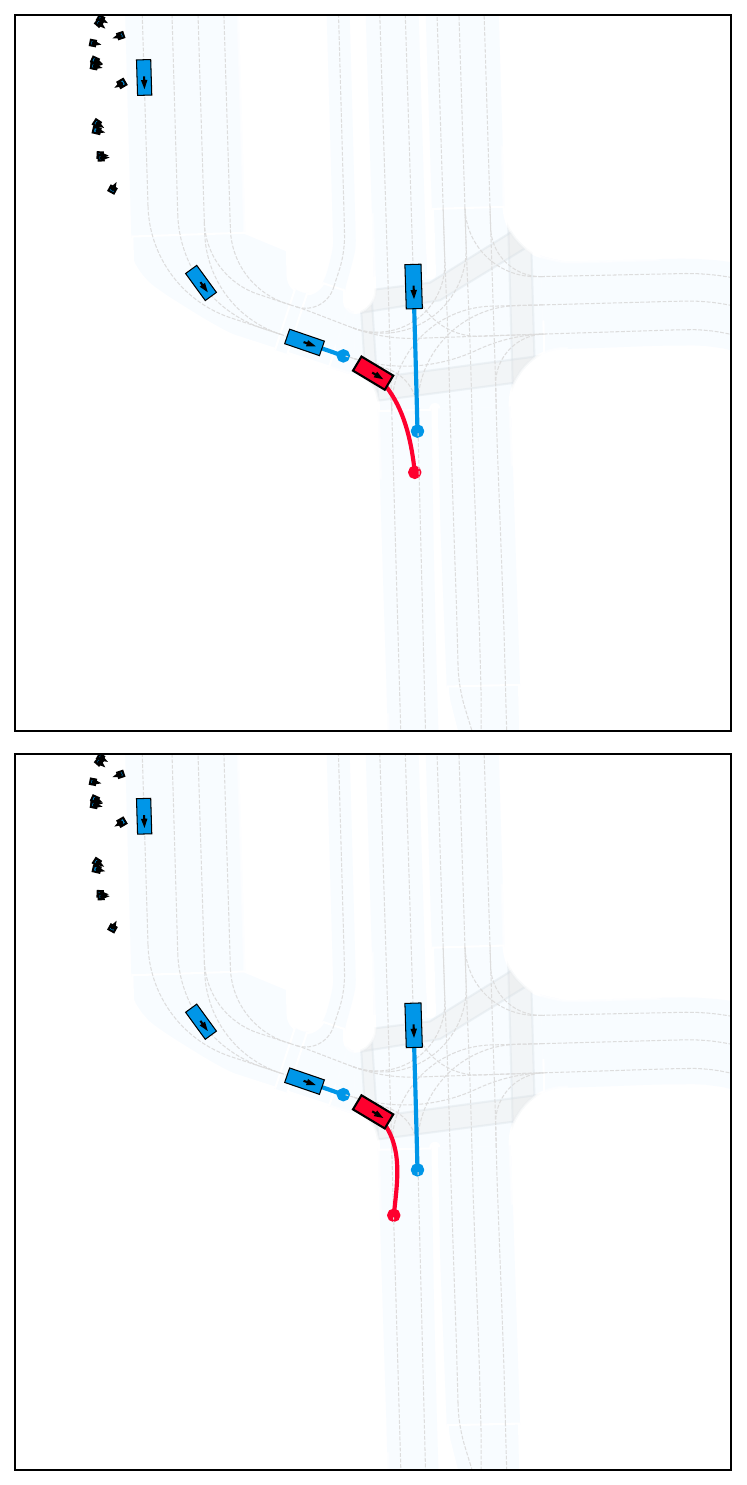}
        \caption{Scenario 4}
    \end{subfigure}
    \begin{subfigure}[b]{0.19\columnwidth}
        \centering
        \includegraphics[width=\textwidth]{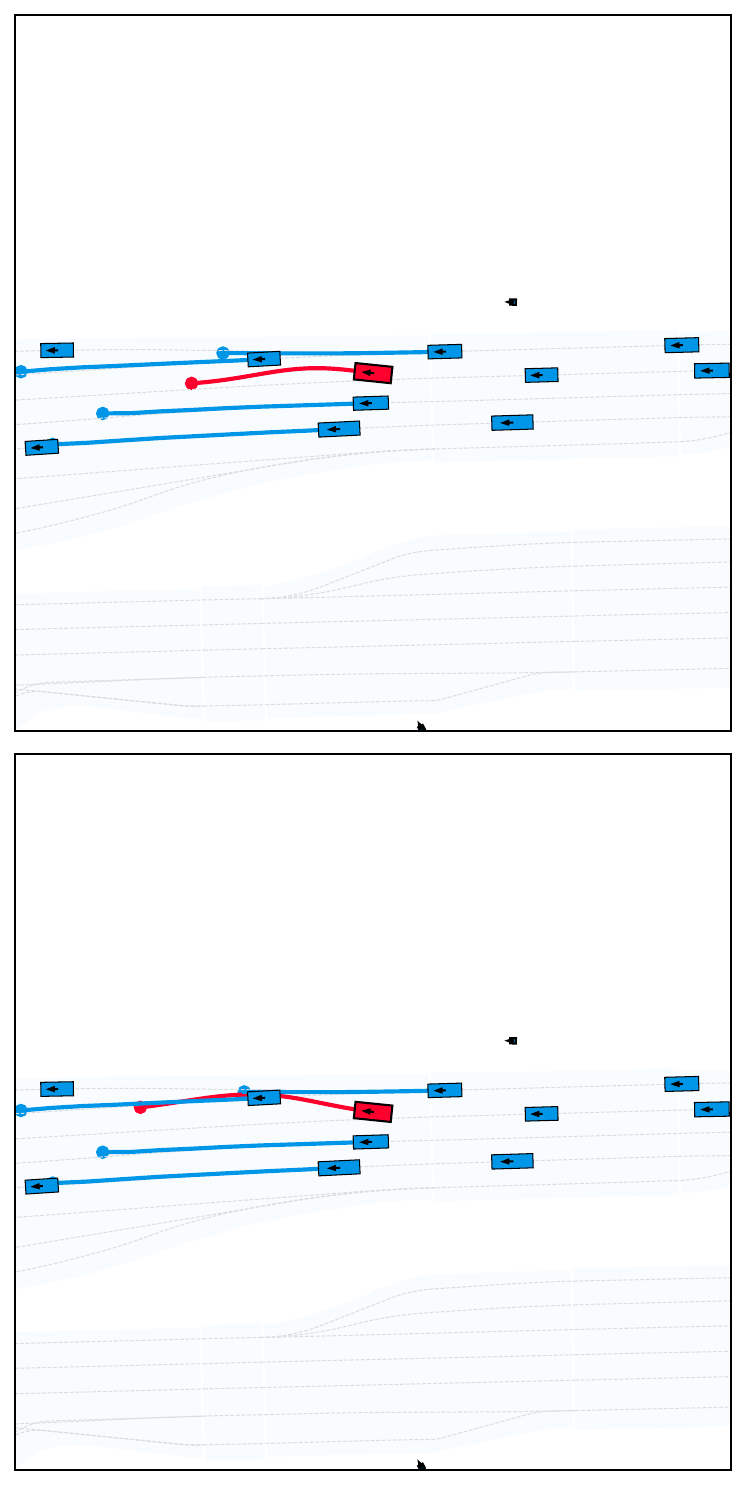}
        \caption{Scenario 5}
    \end{subfigure}
    \caption{\textbf{Top:} highest-scored trajectory per agent from initial distribution before IBR. \textbf{Bottom:} highest-scored trajectory per agent from updated distributions after $k=10$ iterations of IBR. Several scenarios require a lane change to reach the goal or to maintain progress. BIBeR consistently identifies feasible gaps to merge into. For instance, in \textbf{Scenario 1}, the initial trajectory distributions do not permit a safe merge into the adjacent lane. After BIBeR updates the distributions, the ego vehicle infers that the following agent will brake in response to its lane change, enabling a safe maneuver. A similar pattern appears in \textbf{Scenario 2}, where BIBeR anticipates a slight deceleration from the following agent, allowing the ego vehicle to execute a lane change and to accelerate more aggressively compared to the initial trajectories. In \textbf{Scenario 3}, the model handles a particularly challenging lane change: BIBeR accurately predicts the follower’s yielding behavior, making the merge possible, something that a standard prediction-then-planning pipeline would fail to achieve. Comparable behavior can be observed in \textbf{Scenario 4} and \textbf{Scenario 5}, where BIBeR refines the interaction between agents to facilitate safer and more coordinated maneuvers. BIBeR did not cause a collision in any of these scenarios. Ego agent and trajectory are shown in \textcolor{red}{red}, while surrounding agents and corresponding trajectories are shown in \textcolor{blue}{blue}.}
    \label{fig:ibr_qual_res}
\end{figure}

BIBeR updates only the trajectory distributions of the ego vehicle and surrounding agents according to the reward function in Eq.~\ref{eq:reward}. This involves evaluating collisions and distances between all agents, as well as computing the progress and comfort for each ego proposal (note that progress and comfort are computed exclusively for the ego). To reduce computational overhead, we exploit the fact that BIBeR modifies only the trajectory weights, not the trajectories themselves. Consequently, collisions and distances between all trajectories, as well as comfort and progress for all ego proposals, are computed once prior to entering the IBR loop and are subsequently reused to evaluate the reward of each trajectory during the iterations. We provide qualitative examples of ego proposals and traffic agent trajectories before IBR and after IBR in Fig.~\ref{fig:ibr_qual_res}.

\begin{figure}[t]
    \centering
    \begin{subfigure}[t]{0.48\columnwidth}
        \centering
        \includegraphics[width=\linewidth]{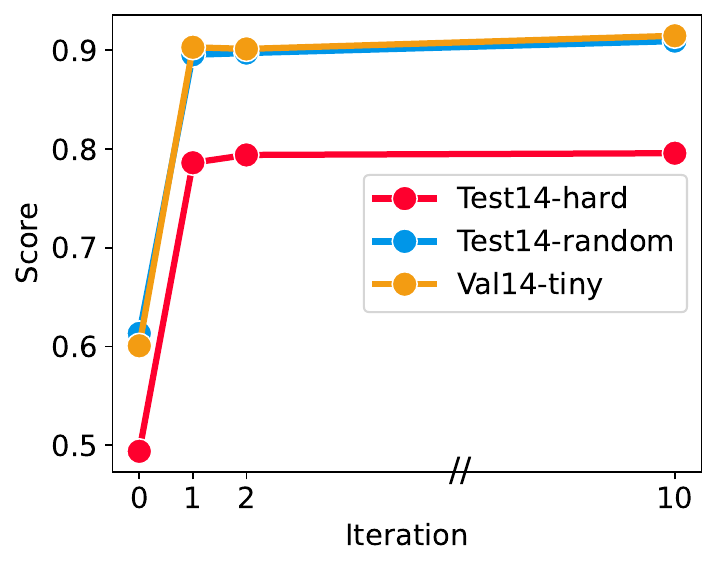}
        \caption{}
        \label{fig:iterations_app}
    \end{subfigure}
    \hfill
    \begin{subfigure}[t]{0.48\columnwidth}
        \centering
        \includegraphics[width=\linewidth]{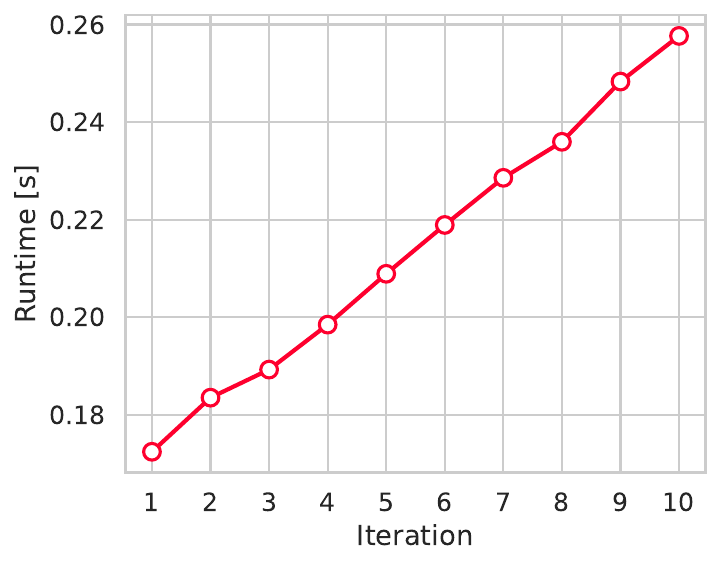}
        \caption{}
        \label{fig:runtime}
    \end{subfigure}
    \caption{Shows \textbf{(a)} performance of BIBeR on different benchmarks across varying numbers of iterations. Val14-tiny is a subset of the Val14 benchmark and \textbf{(b)} runtime of the IBR procedure across different numbers of iterations.}
    \vspace{-0.2cm}
    \label{fig:runtime_combined}
\end{figure}

We conducted additional experiments across more benchmarks using 0, 1, 2, and 10 iterations in Fig.~\ref{fig:iterations_app}. Performance after the second iteration is already comparable to 069 that at the tenth iteration, indicating that two IBR iterations are sufficient to achieve strong performance.

In addition, we analyzed the Test14-hard results and measured the average number of iterations required for convergence. Results show that BIBeR required more iterations to converge in complex scenarios than in simpler ones. For example, convergence was achieved after an average of $1.33$ iterations in stationary-in-traffic scenarios, whereas left-turn scenarios required $2.22$ iterations on average.
\label{sec:suppl:biber-details}

\subsection{Hyperparameters}
\label{subsec:suppl:hyperparameters}
\begin{table}[t]
\centering
\begin{tabular}{lc}
\toprule
\textbf{Hyperparameter} & \textbf{Value} \\
\midrule
Collision cost $u_c$ & $-1.5$\\
Distance cost $u_d$ & $-1.5$\\
Longitudinal progress weight $\alpha$ & $0.19$\\
Lateral progress weight  $\beta$ & $0.1$\\
Comfort weight $w_c$ & $0.15$ \\
Progress weight $w_p$ & $0.9$\\
Number of iterations & $10$ \\
Number of ego proposals $M_e$ & 128\\
Number of modes per agent $M_a$ & 5 \\
Ego order & First\\
Confidence estimation & True\\
Length of proposals & $4.0$ s\\
\bottomrule
\end{tabular}
\caption{Hyperparameter settings of BIBeR we used for our experiments. For BIBeR-CV, we used $M_e = 28$ ego proposals.}
\label{tab:suppl_biber_hps}
\end{table}
Sec.~\ref{sec:biber} introduces some hyperparameters of BIBeR, which are summarized in Tab.~\ref{tab:suppl_biber_hps}. While we did not perform exhaustive hyperparameter optimization, we conducted a series of ablation experiments to assess their influence on performance. We observed that varying the weights and costs associated with progress, comfort, and collision avoidance had only a minor effect on overall performance. However, our weight choices remained within reasonable ratios, for example, comfort was never assigned a higher weight than progress.
In contrast, the agent update order had a substantial impact. 
Alg.\ref{alg:biber} and Eq.\ref{eq:reward} follow a fixed agent update order, if the ego agent is updated first, the other agents tend to adapt to it, whereas if it is updated last, the ego adapts more strongly to the behavior of the others.
When the ego agent was updated last, the system behaved more like a prediction--planning pipeline, which resulted in a significant drop in performance. The number of iterations exhibited only a modest influence: the largest performance gain occurred after the first iteration, with subsequent iterations yielding only marginal improvements.

\subsection{Reward}
\begin{figure}[t]
    \centering
    \begin{subfigure}[b]{0.49\columnwidth}
        \centering
        \includegraphics[width=\textwidth]{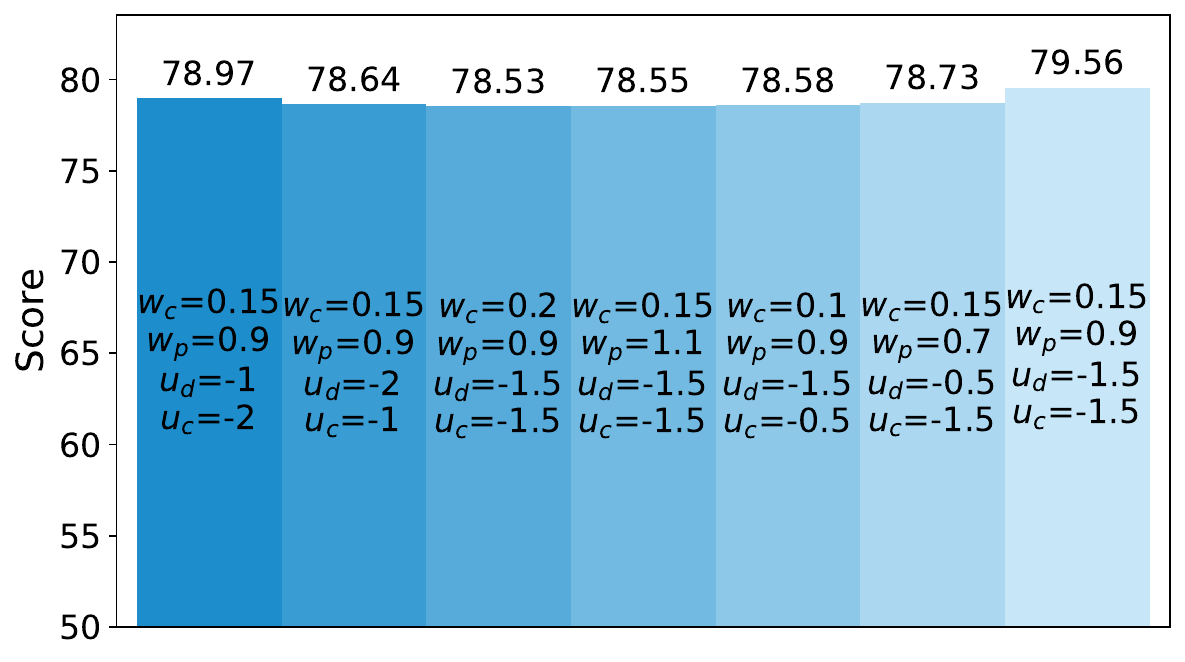}
        \caption{}
        \label{fig:reward_ablation_1}
    \end{subfigure}
    \begin{subfigure}[b]{0.49\columnwidth}
        \centering
        \includegraphics[width=\textwidth]{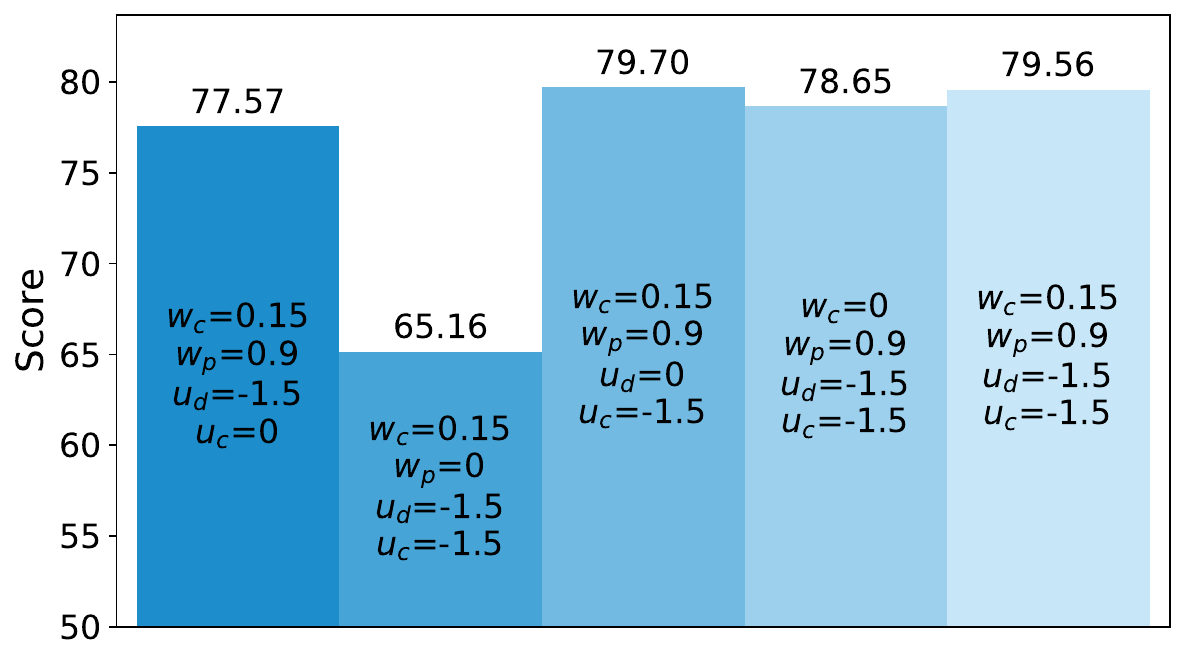}
        \caption{}
        \label{fig:reward_ablation_2}
    \end{subfigure}
    \caption{Performance of BIBeR on Test14-hard for \textbf{(a)} different weightings of the individual reward terms and \textbf{(b)} the removal of entire reward terms. $w_c$ denotes the weight of the comfort term, $w_p$ the weight of the progress term, $u_d$ the penalty for distance violation, and $u_c$ the penalty for a collision.}
    \label{fig:reward_weights_ablation}
\end{figure}

In our reward function, we incorporate three components: distance, progress, and comfort. The distance component is defined in Eq.~\ref{eq:r_distance}. The progress component is composed of two parts: a longitudinal term and a lateral term. The longitudinal term measures progress along the route direction, while the lateral term measures progress relative to the route laterally. These two terms are weighted by $\alpha$ and $\beta$, respectively. For the comfort component, we adopt the comfort reward used in\cite{dauner2023parting}.
We further include additional experiments on the Test14-hard benchmark evaluating different reward weight combinations (Fig.~\ref{fig:reward_ablation_1}) as well as an ablation study of individual reward terms (Fig.~\ref{fig:reward_ablation_2}). The results show that varying the reward weights has only a minor effect on performance, demonstrating the robustness of the reward design. In contrast, removing reward terms leads to a noticeable performance drop, with the progress term contributing the most to the degradation. We additionally conducted an ablation study on the interPlanLC benchmark, where we observed a larger performance degradation when ablating the collision terms, since it contains more interactive scenarios.

\label{subsec:suppl:reward}

\subsection{Runtime}
\label{subsec:suppl:runtime}
Game-theoretic methods are often limited by their substantial computational demands. In Fig.~\ref{fig:runtime}, we report the runtime of BIBeR for different numbers of iterations. The measured runtime covers the IBR and Bayesian confidence estimation components. BIBeR is dependent on the proposal generation and prediction approach and thus potentially highly parallelizable, therefore we excluded them from the runtime analysis. Fig.~\ref{fig:runtime} demonstrates that runtime increases linearly with the number of iterations, although our implementation does not run under the common real-time system constraints of $10\,\text{Hz}$, a more efficient implementation of the algorithm (e.g. in C++ or C) for a real system would be real-time suitable. Our experiments in Tab.~\ref{tab:iterations} and in Fig.~\ref{fig:iterations_app} further indicate that ten iterations are not required to achieve strong performance, in fact, very competitive results are obtained after the first iteration. Additional optimizations may further reduce runtime.

\subsection{Detailed Comparison to SPDM}
\label{subsec:suppl:comparison-spdm}
We present a more detailed comparison between SPDM, BIBeR and BIBeR-CV in Tab.~\ref{tab:suppl-spdm-biber-bibercv-all}.
Across the benchmarks, several consistent trends emerge regarding the relative strengths of BIBeR and BIBeR-CV compared to SPDM. In the highly interactive interPlanLC-SR benchmark, both BIBeR variants achieve particularly strong collision-related scores as well as strong progress, clearly outperforming SPDM. TTC is also much better but just for BIBeR, indicating that it benefits from better agent models significantly more. This indicates that modeling mutual dependencies between agents can simultaneously enhance safety and efficiency when interaction levels are high.
In contrast, on the less interactive benchmarks such as Val14 and Test14-random, BIBeR does not yield measurable improvements over SPDM in either collision avoidance or progress. BIBeR-CV, however, shows a consistent improvement in collision scores, and in the SR setting it achieves noticeably larger gains, particularly in safety-related metrics. Although initially surprising, this result indicates that simpler predictors still perform well under better agent models. 
The Test14-hard benchmark represents an intermediate level of interaction, and the relative performance reflects this: BIBeR improves collision metrics over SPDM, while BIBeR-CV improves both collision rates and progress by a larger margin than on Val14 and Test14-random. This pattern suggests that BIBeR becomes increasingly beneficial as the interaction complexity of the scenario increases. We also report the complementary final score comparison in Tab.~\ref{tab:spdm_biber_r} and in Tab.~\ref{tab:spdm_biber_cv_sr}.

\begin{table}[t]
\resizebox{1\columnwidth}{!}{
\centering
\small
\setlength{\tabcolsep}{4pt}
\begin{tabular}{ll|rrr|rrr|rrr}
\toprule
Benchmark & Subscore & \multicolumn{3}{c|}{CLS-NR $\uparrow$} & \multicolumn{3}{c|}{CLS-R $\uparrow$} & \multicolumn{3}{c}{CLS-SR $\uparrow$} \\
\cmidrule(lr){3-5} \cmidrule(lr){6-8} \cmidrule(lr){9-11}
 & & SPDM & BIBeR & BIBeR-CV & SPDM & BIBeR & BIBeR-CV & SPDM & BIBeR & BIBeR-CV \\
\midrule
\multirow{6}{*}{Test14-hard} 
 & Comfort & 85.29 & 83.82 & \textbf{86.76} & 84.93 & 83.82 & \textbf{84.19} & 84.93 & \textbf{87.13} & 84.19 \\
 & MP      & \textbf{95.59} & 91.18 & \textbf{95.59} & \textbf{94.85} & 92.28 & \textbf{94.85} & 88.24 & 88.24 & \textbf{90.07} \\
 & EP      & \textbf{79.08} & 73.95 & 79.26 & \textbf{77.96} & 73.50 & 77.96 & 71.93 & 70.57 & \textbf{74.93} \\
 & No-Col. & 93.38 & 93.93 & \textbf{96.14} & 97.24 & 97.43 & \textbf{98.71} & \textbf{97.79} & \textbf{97.79} & 97.05 \\
 & TTC     & 77.57 & 79.04 & \textbf{83.46} & 88.60 & \textbf{90.81} & 89.71 & 88.60 & \textbf{90.07} & 87.13 \\
\midrule
\multirow{5}{*}{Test14-random} 
 & Comfort & \textbf{96.06} & 92.86 & 93.93 & 91.76 & 92.14 & \textbf{94.29} & \textbf{93.21} & 90.36 & 92.50 \\
 & MP      & \textbf{99.28} & 97.86 & 98.57 & \textbf{98.57} & 97.14 & 97.86 & \textbf{97.50} & 95.71 & 96.78 \\
 & EP      & \textbf{92.66} & 89.34 & 91.99 & \textbf{90.41} & 87.10 & 89.60 & 89.51 & 87.10 & \textbf{90.55} \\
 & No-Col. & 97.49 & \textbf{99.29} & 98.93 & 99.10 & 98.93 & \textbf{99.29} & 97.14 & 97.50 & \textbf{98.21} \\
 & TTC     & 92.11 & 92.50 & \textbf{95.71} & 94.62 & 95.00 & \textbf{97.86} & 92.14 & 90.36 & \textbf{93.57} \\
\midrule
\multirow{5}{*}{Val14} 
 & Comfort & \textbf{93.82} & 91.68 & 93.82 & 92.03 & 89.53 & \textbf{92.22} & 93.92 & 92.93 & \textbf{94.27} \\
 & MP      & \textbf{99.28} & 98.21 & 98.92 & \textbf{99.37} & 98.03 & 98.93 & \textbf{96.78} & 96.15 & \textbf{96.78} \\
 & EP      & \textbf{91.95} & 89.29 & 91.95 & \textbf{90.33} & 87.40 & 90.21 & 87.29 & 84.63 & \textbf{87.87} \\
 & No-Col. & 97.40 & \textbf{98.52} & 97.85 & 98.43 & 98.52 & \textbf{98.88} & 97.94 & 97.67 & \textbf{98.21} \\
 & TTC     & \textbf{92.03} & 89.89 & 91.58 & 94.53 & 91.86 & \textbf{94.19} & 91.32 & 91.95 & \textbf{92.39} \\
\midrule
\multirow{5}{*}{interPlanLC} 
 & Comfort & N/A & N/A & N/A & \textbf{90.00} & 83.33 & 79.99 & \textbf{93.33} & 73.33 & 80.00 \\
 & MP      & N/A & N/A & N/A & \textbf{100.00} & \textbf{100.00} & \textbf{100.00} & \textbf{96.67} & 93.33 & 96.66 \\
 & EP      & N/A & N/A & N/A & 64.08 & 65.94 & \textbf{66.23} & 51.95 & 50.29 & \textbf{54.09} \\
 & No-Col. & N/A & N/A & N/A & \textbf{100.00} & \textbf{100.00} & \textbf{100.00} & 93.33 & \textbf{96.67} & 90.00 \\
 & TTC     & N/A & N/A & N/A & 96.67 & \textbf{100.00} & \textbf{100.00} & 66.67 & \textbf{86.67} & 66.66 \\
\midrule
\multirow{5}{*}{interPlan} 
 & Comfort & N/A & N/A & N/A & \textbf{95.00} & 92.50 & 91.25 & N/A & N/A & N/A \\
 & MP      & N/A & N/A & N/A & \textbf{87.50} & \textbf{87.50} & \textbf{87.50} & N/A & N/A & N/A \\
 & EP      & N/A & N/A & N/A & 59.65 & 60.30 & \textbf{61.51} & N/A & N/A & N/A \\
 & No-Col. & N/A & N/A & N/A & \textbf{91.25} & 90.00 & \textbf{91.25} & N/A & N/A & N/A \\
 & TTC     & N/A & N/A & N/A & \textbf{90.00} & 87.50 & \textbf{90.00} & N/A & N/A & N/A \\
\bottomrule
\end{tabular}
}
\caption{Compares the performance of SPDM, BIBeR, and BIBeR-CV across multiple benchmarks (Test14-hard, Test14-random, Val14, interPlanLC, and interPlan) in different simulation settings: non-reactive (CLS-NR), reactive (CLS-R), and SMART-reactive (CLS-SR). All five subscores for each benchmark are shown. Comfort, MP (Making Progress), EP (Ego Progress), No-Collision, and TTC (Time-to-Collision). Higher values indicate better performance. The results highlight the relative strengths of the three planners across different settings. NR setting is not available for interPlan and interPlanLC and SR setting is not available for interPlan. Bold values marking the best subscore performance in each benchmark.}
\label{tab:suppl-spdm-biber-bibercv-all}
\end{table}
\begin{table}[t]
    \centering
    \begin{tabular}{lcccc}
        \toprule
        \multirow{2}{*}{Method}
        & \multicolumn{4}{c}{CLS-R $\uparrow$} \\
        \cmidrule(lr){2-5}
        & Val14 & Test14-hard & Test14-random & interPlanLC \\
        \midrule
        SPDM~\cite{distelzweig2025perfectprediction}
        & \textbf{92.72} & \textbf{80.75} & \textbf{92.54} & \textbf{84.07} \\
        \textbf{BIBeR} (Ours) 
        & 89.79 & 79.57 & 90.92 & 81.17 \\
        \bottomrule
    \end{tabular}
    \caption{Comparison between SPDM and BIBeR on the nuPlan and interPlan benchmarks. We report closed-loop reactive (CLS-R) performance on Val14, Test14-hard, Test14-random, and interPlanLC.}
    \vspace{-0.2cm}
    \label{tab:spdm_biber_r}
\end{table}

\begin{table}[t]
    \centering
    \begin{tabular}{lcccc}
        \toprule
        \multirow{2}{*}{Method}
        & \multicolumn{4}{c}{CLS-SR $\uparrow$} \\
        \cmidrule(lr){2-5}
        & Val14
        & Test14-hard
        & Test14-random
        & interPlanLC \\
        \midrule
        SPDM~\cite{distelzweig2025perfectprediction} 
            & 88.53 & 75.01 & 89.50 & 63.40 \\
        \textbf{BIBeR-CV} (Ours) 
            & \textbf{89.57} & \textbf{76.25} & \textbf{90.22} & \textbf{66.04} \\
        \bottomrule
    \end{tabular}
    \caption{SMART Reactive (CLS-SR) evaluation on nuPlan and interPlan benchmarks, comparing SPDM and BIBeR-CV. Results are reported on Val14, Test14-hard, Test14-random, and interPlanLC.}
    \vspace{-0.2cm}
    \label{tab:spdm_biber_cv_sr}
\end{table}

\subsection{Additional Results}
\label{subsec:suppl:additional-res-biber}
\begin{figure}
    \center
    \includegraphics[width=1\columnwidth]{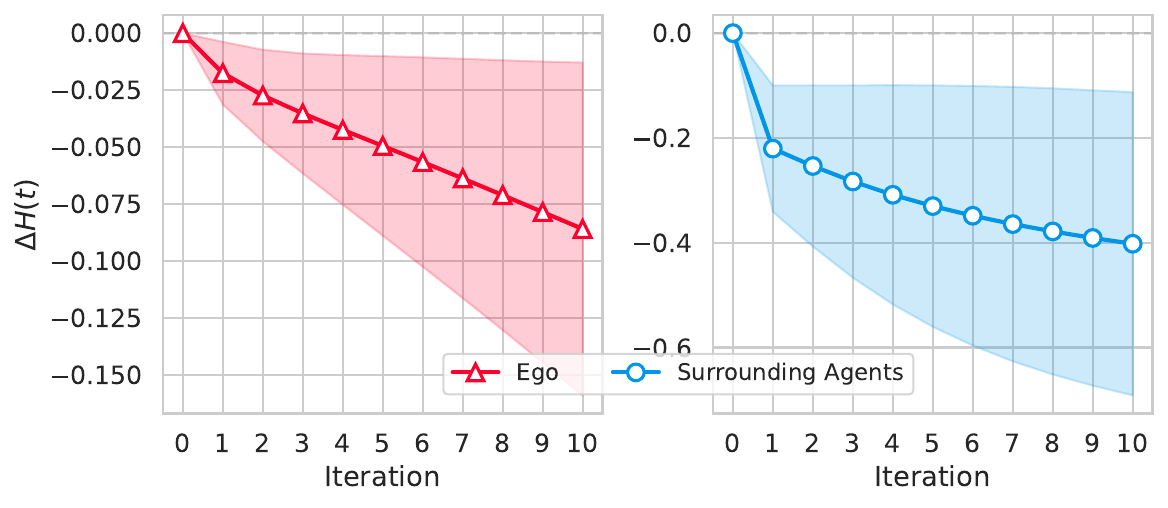}
    \caption{Relative entropy over iterations on the Test14-hard reactive benchmark using BIBeR with LAformer as the prediction model. \textbf{Left:} mean and standard deviation of the relative entropy of the ego trajectory distribution across iterations. \textbf{Right:} mean and standard deviation of the relative entropy of trajectory distributions from surrounding agents across iterations. Relative entropy is computed w.r.t. the initial distribution entropy.}
    \vspace{-0.5cm}
    \label{fig:entropy}
\end{figure}
To demonstrate that BIBeR converges to a specific scenario, i.e., a particular mode, we report the relative entropy of both the agents’ and the ego vehicle’s trajectory distributions over iterations and multiple scenarios in Fig.~\ref{fig:entropy}. Specifically, we measure the relative entropy with respect to the distributions before the first iteration. Fig.~\ref{fig:entropy} reveals a clear trend: the entropy of both ego and surrounding agents’ trajectory distributions decreases over iterations. This indicates that the probability mass gradually concentrates on a specific mode, suggesting that the updates do not oscillate between different modes but instead converge to a consistent outcome.
\section{Prediction Models}
\label{sec:suppl:prediction-models}
\begin{table}[t]
\centering
\begin{subtable}[t]{0.45\columnwidth}
\centering
\resizebox{\linewidth}{!}{
\begin{tabular}{lc}
\toprule
\textbf{Hyperparameter} & \textbf{Value} \\
\midrule
Size of latent vector ($z_{size}$) & 2 \\
Number of top-$k$ lane segments & 2 \\
Batch size & 8 \\
Lane loss weight & 0.9 \\
Learning rate & 0.001 \\
Learning rate decay & 0.01 \\
Number of training epochs & 5 \\
Size of encodings ($h_{size}$) & 64 \\
Size of feature vector & 32 \\
Hidden dropout probability & 0.1 \\
Initializer range & 0.02 \\
Number of future frames & 12 \\
Number of history frames & 4 \\
Future trajectory length (s) & 6 \\
History trajectory length (s) & 2 \\
Number of output modes & 5 \\
\bottomrule
\end{tabular}
}
\caption{Hyperparameter settings of LAformer~\cite{liu2024laformer} used for training.}
\label{tab:suppl_laformer_hps}
\end{subtable}
\hfill
\begin{subtable}[t]{0.45\columnwidth}
\centering
\resizebox{\linewidth}{!}{
\begin{tabular}{lc}
\toprule
\textbf{Hyperparameter} & \textbf{Value} \\
\midrule
Encoder layers & 6 \\
Decoder levels & 0 \\
Learning rate & 0.0001 \\
Agent encoding size & 256 \\
Ego encoding size & 256 \\
Map encoding size & 256 \\
Encoder heads & 8 \\
Encoder dropout rate & 0.1 \\
Mode query embedding size & 256 \\
Agent query embedding size & 256 \\
Decoder heads & 8 \\
Decoder dropout rate & 0.1 \\
Number of output modes & 5 \\
Batch size &  64 \\
Number of training epochs & 33\\
\bottomrule
\end{tabular}
}
\caption{Hyperparameter settings of Joint Predictor~\cite{huang2023gameformer} used for training.}
\vspace{-0.2cm}
\label{tab:suppl_jp_hps}
\end{subtable}
\caption{Hyperparameter settings for LAformer and Joint Predictor used in our experiments.}
\label{tab:suppl_hyperparams_combined}
\end{table}

\subsection{LAformer}
\label{subsec:suppl:laformer}
LAformer~\cite{liu2024laformer} is a state-of-the-art, attention-based marginal trajectory predictor that leverages a lane-aware estimation module and outputs predictions for a single target agent.

\subsubsection{Architecture Overview}
\label{subsubsec:suppl:laformer-arch}
Both agent histories and lane centerlines are represented as vectors. Agent $i$'s past trajectory $X_i^{-t_h:0}$ is encoded as a sequence of sparse vectors $A_i^{-t_h:0} = \{v_i^{-t_h}, \dots, v_i^0\}$, where $v_i^t = [d_{i,s}^t, d_{i,e}^t, a_i]$ contains start/end positions and agent attributes. Lane centerlines are similarly sliced into segments $C_i^{1:N} = \{v_i^1, \dots, v_i^N\}$. Agent motion and scene context is encoded via a Global Interaction Graph (GIG) using MLP and GRU layers, producing encodings $h_i$ for trajectories and $c_j$ for lane segments. Cross-attention fuses these encodings:
\begin{align}
    h_i &\gets h_i + \text{CrossAtt}(h_i, c_j), \quad j \in \{1,\dots,N_\text{lane}\},\\
    c_j &\gets c_j + \text{CrossAtt}(c_j, h_i), \quad i \in \{1,\dots,N_\text{traj}\}.
\end{align}
Self-attention further models agent interactions:
\begin{align}
    h_i &\gets \text{Concat}[h_i, c_j], \quad j \in \{1,\dots,N_\text{lane}\}, \\
    h_i &\gets h_i + \text{SelfAtt}(h_i), \quad i \in \{1,\dots,N_\text{traj}\}.
\end{align}
Future agent motion is aligned with lane segments using attention. Lane probabilities are computed as
\begin{equation}
    \hat{s}_{j,t} = \frac{\exp(\phi(h_i, c_j, A_{i,j}))}{\sum_{n=1}^{N_\text{lane}} \exp(\phi(h_i, c_n, A_{i,n}))}, \quad t \in \{1,\dots,t_f\},
\end{equation}
where $A_{i,j} = \text{softmax}(Q K^\top / \sqrt{d_k}) V$ and $\phi$ is a two-layer MLP. Top-$k$ lanes with highest scores are selected as candidates and concatenated over time: $C = \text{Concat}\{c_{1:k}, \hat{s}_{1:k}\}_{t=1}^{t_f}$. Cross-attention between $h_i$ and candidate lanes produces lane-aware encodings $h_{i,\text{att}}$.
A Laplacian MDN decoder predicts future trajectories conditioned on $h_i$ and $h_{i,\text{att}}$, with latent vector $z \sim \mathcal{N}(0,I)$ for diversity. Predicted trajectories follow
\begin{equation}
    p(Y|h_i, h_{i,\text{att}}, z) = \sum_{m=1}^{M} \pi_m \text{Laplace}(\mu_m, b_m), \quad \sum_{m=1}^{M} \pi_m = 1.
\end{equation}
In the original work, a second training step is applied for refinement, which we omitted in our work. For further details, we refer to the original work~\cite{liu2024laformer}.

\subsubsection{Training Settings}
\label{subsubsec:suppl:laformer-training}
The open-source LAformer~\cite{liu2024laformer} prediction model was initially trained on the nuScenes dataset, which is recorded at $2\,\text{Hz}$. In our work, we retrain LAformer from scratch on the nuPlan dataset, originally sampled at $10\,\text{Hz}$. To remain consistent with the original training setup, we downsample the nuPlan data to $2\,\text{Hz}$. As in the original configuration, we use a $2\,\text{s}$ history, corresponding to $4$ past positions and the model predicts a $4\,\text{s}$ future trajectory, i.e., $8$ positions at $2\,\text{Hz}$, which we then interpolate to $40$ positions to match the $10\,\text{Hz}$ frequency. We sample 150,000 scenarios from the nuPlan training split for model training. The hyperparameters used to train LAformer are summarized in Tab.~\ref{tab:suppl_laformer_hps}.
The model is trained using a single NVIDIA RTX A6000 GPU.

\subsubsection{Qualitative Results}
\label{subsubsec:suppl:laformer-qualitative}
\begin{figure}[h]
    \centering
    \begin{subfigure}[b]{0.49\columnwidth}
        \centering
        \includegraphics[width=\textwidth]{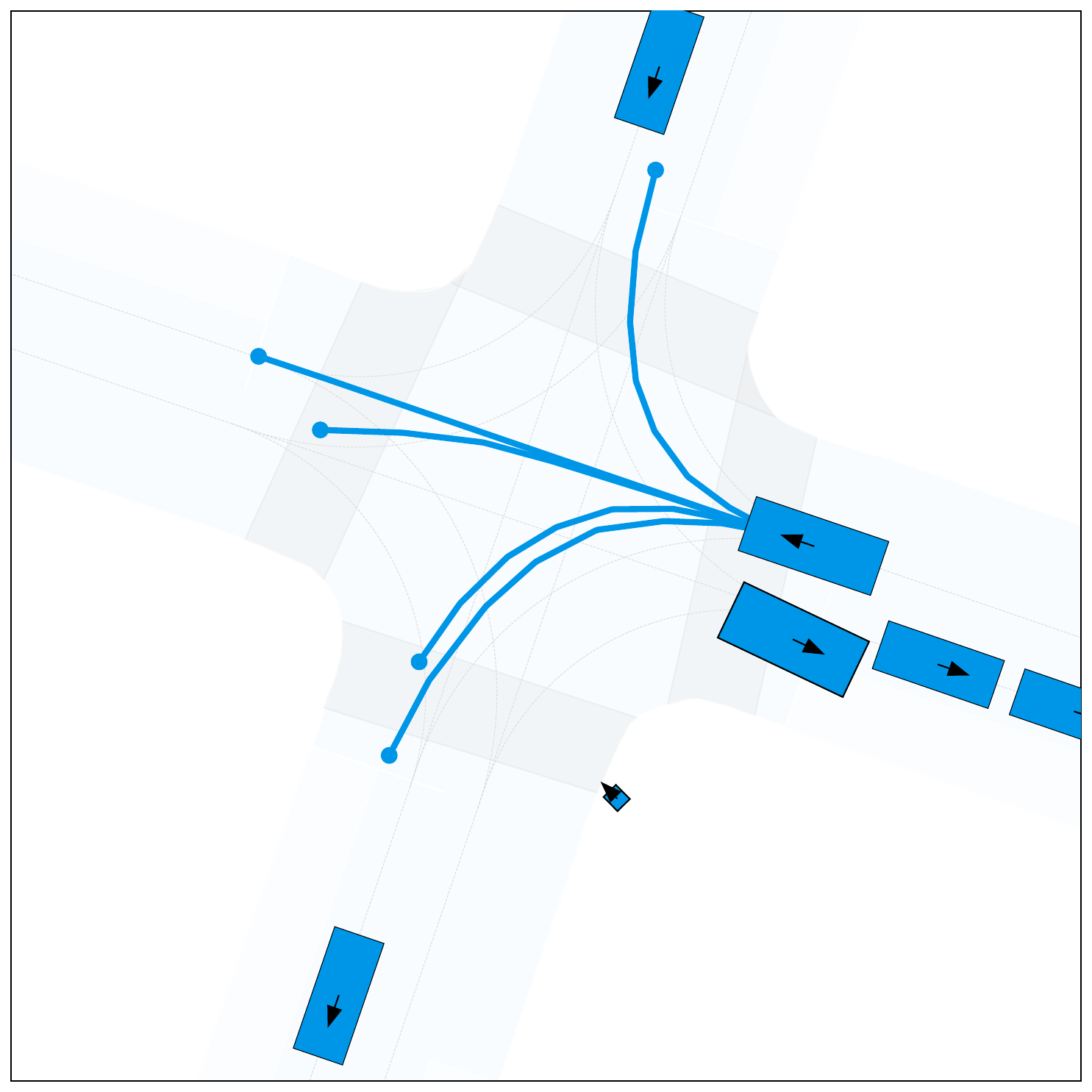}
    \end{subfigure}
    \hfill
    \begin{subfigure}[b]{0.49\columnwidth}
        \centering
        \includegraphics[width=\textwidth]{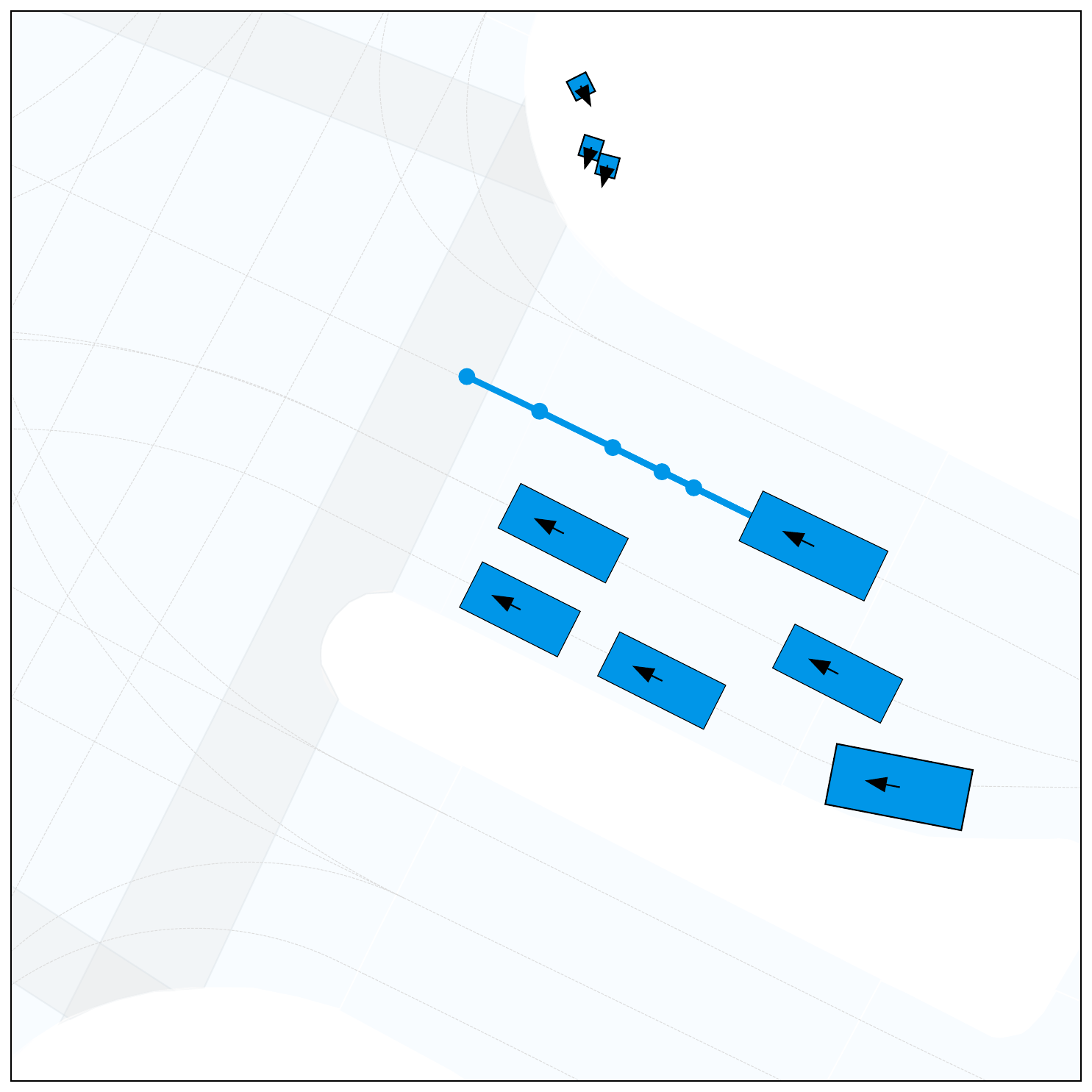}
    \end{subfigure}
    % Gesamte Figure Caption
    \caption{Two example scenarios from the Val14 dataset, each showing five LAformer predictions for a single agent, illustrating the model’s ability to capture multiple plausible future paths.}
    \label{fig:laformer_preds}
\end{figure}

We provide qualitative results of LAformer on two example scenarios in Fig.~\ref{fig:laformer_preds}, demonstrating its ability to generate multi-modal forecasts.

\subsection{Joint Predictor}
\label{subsec:suppl:joint-predictor}
We use the Joint Predictor from \cite{huang2023gameformer}, a non-autoregressive Transformer encoder–decoder model that produces scene-consistent trajectories for all agents simultaneously. In our experiments, we include this joint predictor to demonstrate that marginal prediction is better suited for BIBeR: joint prediction implicitly commits to a specific mode for surrounding agents, which in turn forces BIBeR into that same mode, whereas BIBeR requires mode indifference to identify the most appropriate behavior.

\subsubsection{Architecture Overview}
\label{subsubsec:suppl:joint-arch}
The model processes the motion history of all agents in the scene and a vectorized representation of the surrounding map. For each agent \(i\), the sequence of past positions \(X_i^{-t_h:0}\) is encoded by an LSTM network, which produces a latent feature tensor containing the past motion features of all agents. The resulting representation is $A_p \in \mathbb{R}^{N \times D}$, where $N$ denotes the number of agents and $D$ the dimensionality of the hidden features.

The road layout around every agent is represented through a collection of local polylines. Each agent is associated with $N_m$ relevant map elements such as lane centerlines or crosswalks, and each of these elements is described by $N_p$ waypoints of dimension $d_p$. These map elements are organized into a tensor $M \in \mathbb{R}^{N \times N_m \times N_p \times d_p}$. An MLP is applied to each waypoint in order to obtain embedded map features. This produces a tensor of the same size but with feature dimension \(D\). To yield a more compact set of tokens per map element, a max pooling operation is carried out over the waypoint dimension. After this pooling, every agent is associated with a reduced set of map features $M_r \in \mathbb{R}^{N \times N_{mr} \times D}$ , where $N_{mr}$ indicates the number of aggregated map elements.

For each agent the encoded motion features are combined with the corresponding map features by concatenation, which results in a context representation that contains both agent information and local map cues. A Transformer encoder is then applied independently to the context of every agent in order to model interactions among the elements within its local scene. Processing all agents produces a scene-level context tensor $C_s \in \mathbb{R}^{N \times (N + N_{mr}) \times D}$. To account for the intrinsic uncertainty of the future, the model introduces a set of learnable modality embeddings. These embeddings are merged with the encoded historical features of each agent, which creates modality-aware query tokens. The resulting tensor serves as the query input to a cross-attention module, while the scene context acts as key and value. Attention is applied along the modality axis for each agent and yields a set of representations $Z \in \mathbb{R}^{N \times M \times D}$. Two prediction heads based on MLPs decode these features into multi-modal futures. The first head outputs Gaussian parameters for every modality and future timestep, expressed as $Y \in \mathbb{R}^{N \times M \times t_f \times 4}$, corresponding to the mean values \(\mu_x\) and \(\mu_y\) and the logarithms of the standard deviations \(\log \sigma_x\) and \(\log \sigma_y\). A second MLP produces a score for each predicted trajectory through $P \in \mathbb{R}^{N \times M \times 1}$.

In contrast to the original formulation, we omit additional interaction decoder stages and refer to the original work~\cite{huang2023gameformer} for further details.

\subsubsection{Training Settings}
\label{subsubsec:suppl:joint-training}
We train the joint predictor on the same set of 150,000 scenarios from the nuPlan training split as also used for training LAformer.
The hyperparameters used to train the joint predictor are presented in Tab.~\ref{tab:suppl_jp_hps}. The model is trained using a single NVIDIA RTX A6000 GPU.

\subsubsection{Qualitative Results}
\label{subsubsec:suppl:joint-qualitative}
\begin{figure}[h]
    \centering
    \begin{subfigure}[b]{0.49\columnwidth}
        \centering
        \includegraphics[width=\textwidth]{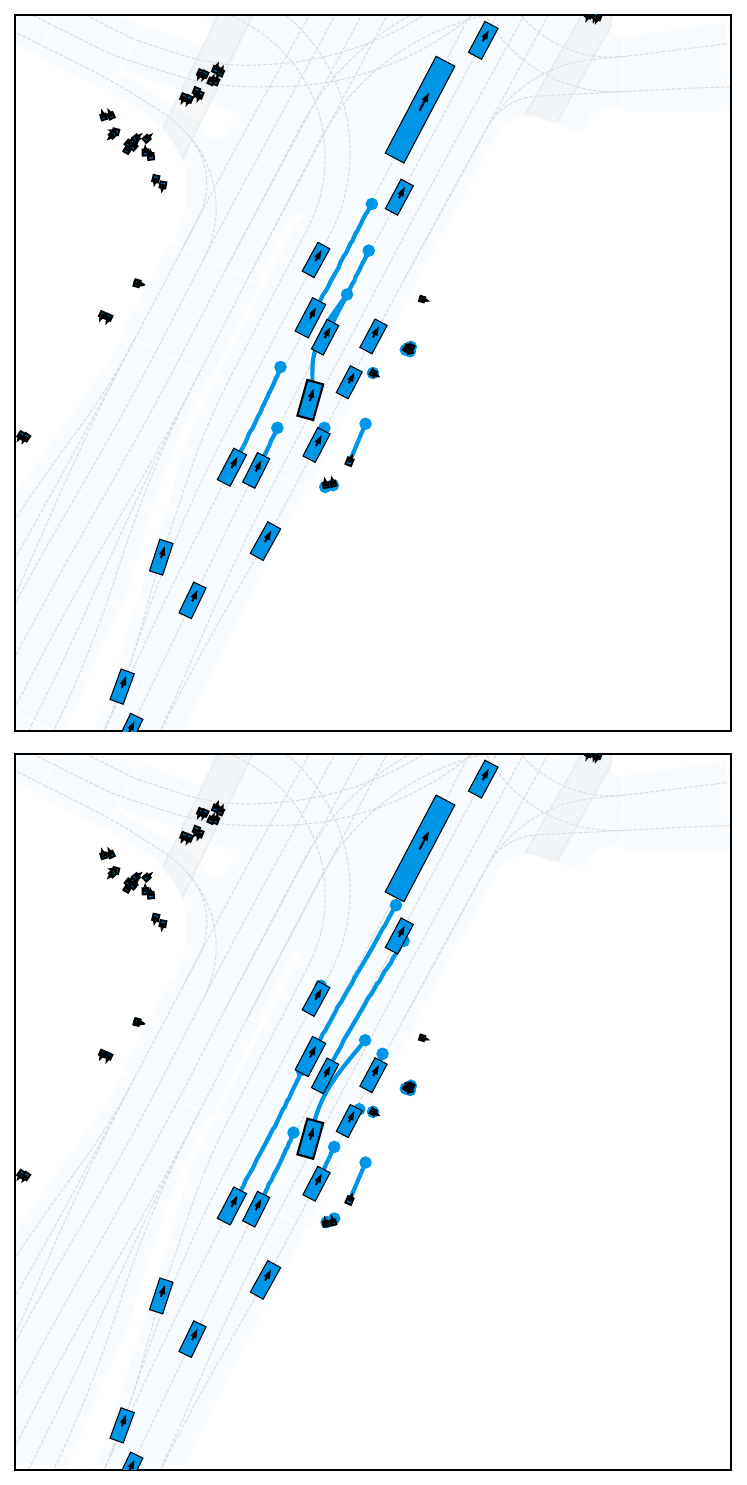}
        \caption{Scenario 1}
    \end{subfigure}
    \hfill
    \begin{subfigure}[b]{0.49\columnwidth}
        \centering
        \includegraphics[width=\textwidth]{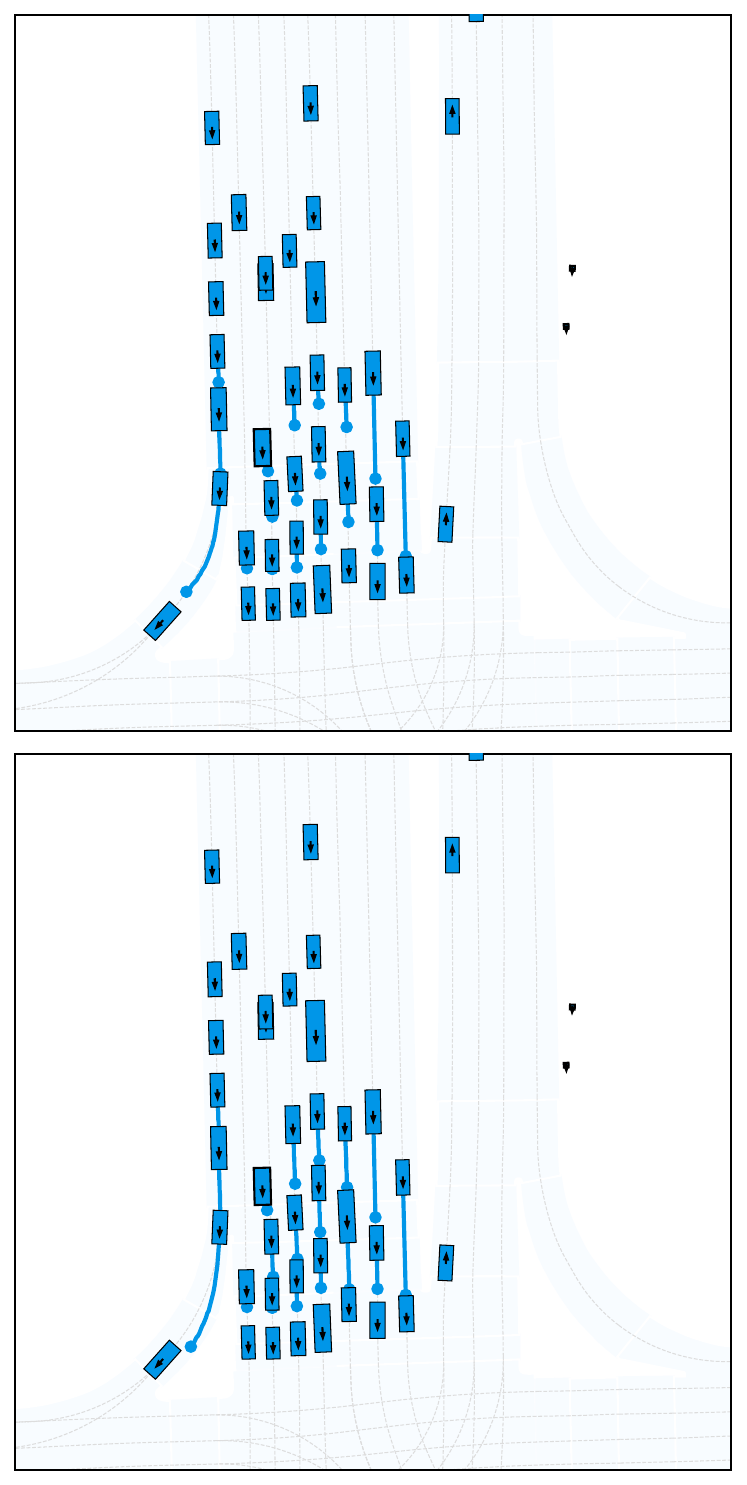}
        \caption{Scenario 2}
    \end{subfigure}
    \caption{Two example scenarios from the Val14 dataset, each column showing two modes of the joint predictor. These illustrate the model’s ability to generate scene-consistent future trajectories. \textbf{Scenario 1:} Two potential different scenarios are predicted: \textbf{Top:} One agent changes lanes while the following agent brakes. \textbf{Bottom:} One agent avoids an occluded gap between two vehicles, prompting the following agent to brake. \textbf{Scenario 2:} several agents approach a red traffic light and decelerate accordingly, with each mode exhibiting different strong decelerations.}
    \label{fig:jp_preds}
\end{figure}

We provide two qualitative results of the joint predictor in Fig.~\ref{fig:jp_preds}, demonstrating its ability to generate scene-consistent future forecasts.
\section{Sampling-based Planner}
\label{sec:suppl:sampling-planners}

\subsection{SPDM}
\label{subsec:suppl:spdm}
SPDM~\cite{distelzweig2025perfectprediction} builds on PDM~\cite{dauner2023parting}. Both planners follow a proposal-selection paradigm: they first generate a diverse set of candidate trajectories and subsequently select the one with the highest score according to a driving-aligned cost function.
SPDM begins by predicting the future motions of surrounding agents using a constant-velocity model. It then extracts the ego vehicle’s current lane and adjacent lanes, explicitly excluding lanes in the opposite driving direction. Similar as in PDM, SPDM generates longitudinal proposals for the current lane using the Intelligent Driver Model (IDM)~\cite{treiber2000congested} with five distinct target-speeds, corresponding to $\{20, 40, 60, 80, 100\}\%$  of the current speed limit.

For each adjacent lane, SPDM first samples a set of points at different longitudinal offsets and fits splines through these points in order to obtain a continuous parameterization of the possible lateral maneuvers. Along each of these splines, the longitudinal evolution is generated using IDM with the same collection of target speed factors as in the main lane. When IDM is applied on the splines of neighboring lanes, only the leading vehicle identified at the initial prediction step is taken into account. This design avoids overly cautious behavior, since considering all approaching vehicles on an adjacent lane would remove virtually all lane change options in high traffic density. Instead, the method relies on the observation that an ego vehicle is able to influence the motion of vehicles behind it through its own actions, while it cannot directly affect the behavior of vehicles in front of it. By fixing the leading vehicle for the entire rollout, the maneuvers remain feasible and ensure safety to the vehicle that cannot be influenced.

\begin{figure}[h]
    \centering
    \begin{subfigure}[b]{0.49\columnwidth}
        \centering
        \includegraphics[width=\textwidth]{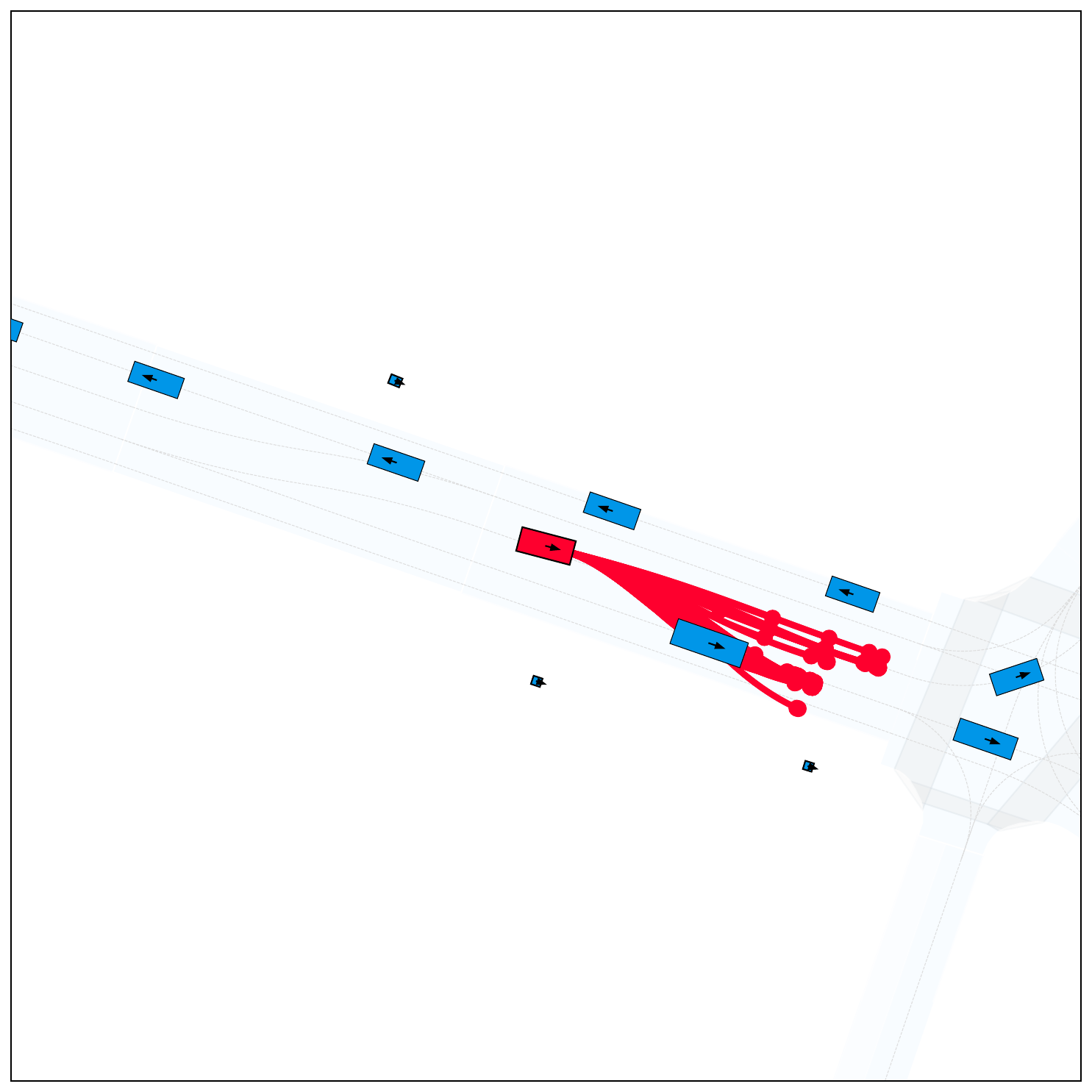}
    \end{subfigure}
    \hfill
    \begin{subfigure}[b]{0.49\columnwidth}
        \centering
        \includegraphics[width=\textwidth]{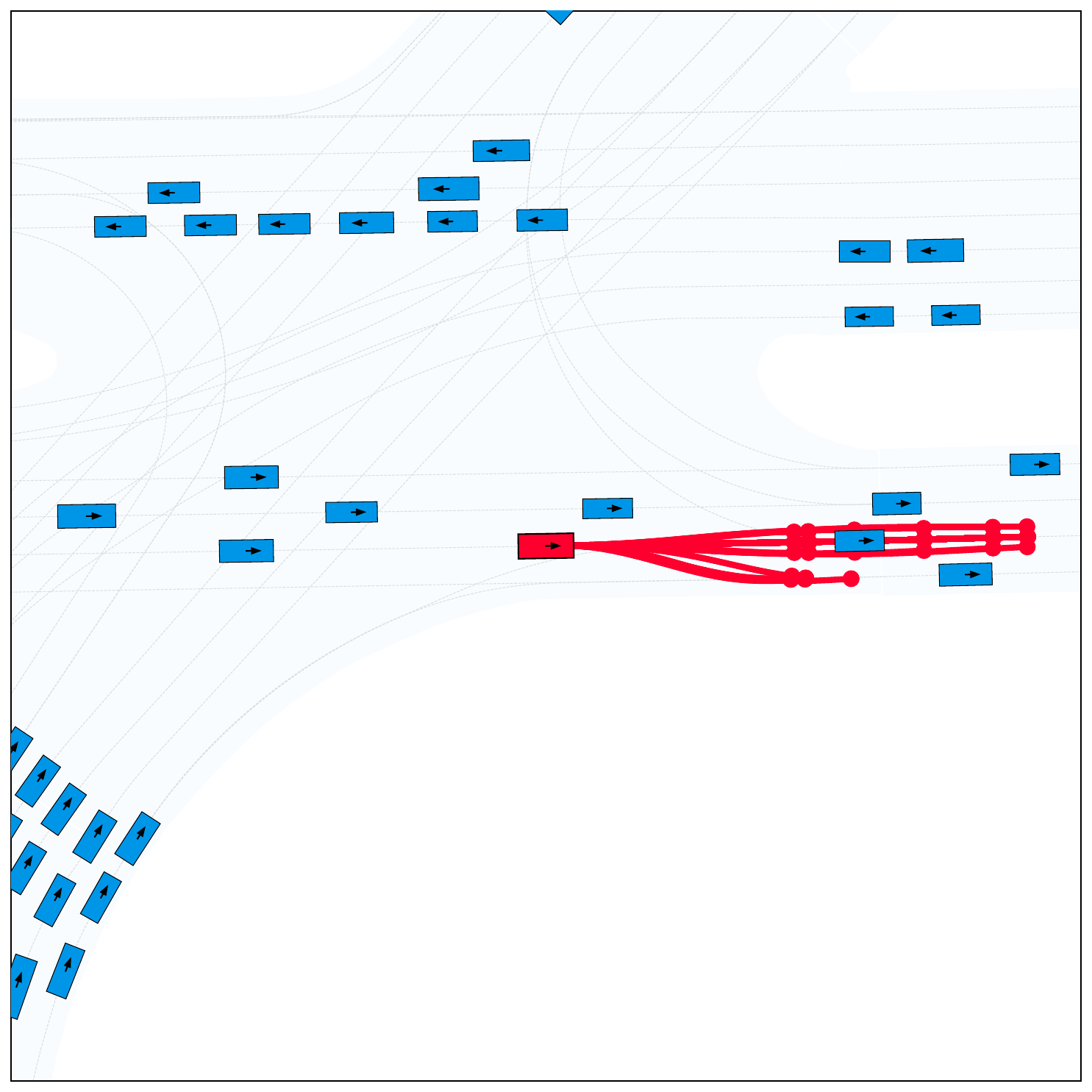}
    \end{subfigure}
    \caption{Two example scenarios from the Val14 dataset, illustrating trajectory proposals generated by SPDM.}
    \label{fig:spdm_proposals}
\end{figure}

All proposals are then simulated using a Linear Quadratic Regulator (LQR) controller and propagated with a kinematic bicycle model. Finally, each simulated trajectory is evaluated using a scoring function closely aligned with the nuPlan evaluation metric (Eq.~\ref{eq:nuplan_score}). The trajectory with the highest score is selected as the final ego plan. In Fig.~\ref{fig:spdm_proposals}, we provide two qualitative results of the SPDM proposals.
\section{Benchmarks}
\label{sec:suppl:benchmarks}
\subsection{Evaluation Metrics}
\label{subsec:suppl:evaluation-metrics}
The overall nuPlan~\cite{caesar2022nuplan} score consists of several metrics (Tab.~\ref{tab:suppl_metric_weights}) that evaluate different aspects of driving performance. 

Drivable Area Compliance (DAC) evaluates whether the ego vehicle remains within drivable area. The metric is binary: $\mathrm{score}_{DAC} = 1$ if the ego stays within the drivable area, and $\mathrm{score}_{DAC} = 0$ otherwise.

Driving Direction Compliance (DDC) penalizes the planner when the ego vehicle enters oncoming traffic. The score is set to $\mathrm{score}_{DDC} = 0$ if the ego drives more than $6\,\mathrm{m}$ in oncoming traffic, $\mathrm{score}_{DDC} = 0.5$ if it drives more than $2\,\mathrm{m}$ but less than $6\,\mathrm{m}$, and $\mathrm{score}_{DDC} = 1$ otherwise.
Making Progress (MP) is based on the ego progress score $\mathrm{score}_{EP}$. If the planner achieves less than $20\%$ of the expert’s progress, the result is $\mathrm{score}_{MP} = 0$, otherwise $\mathrm{score}_{MP} = 1$.

Time to collision refers to the time it would take for the ego vehicle and another object to collide if both continue with their current speed and heading. It is computed only for objects located ahead of the ego, for cross-traffic vehicles, and for lateral objects when the ego is performing a lane change or is navigating through an intersection. The metric yields $\mathrm{score}_{TTC} = 1$ if $\mathrm{TTC} > 0.95\,\mathrm{s}$, and $\mathrm{score}_{TTC} = 0$ otherwise.

Ego Progress (EP) measures the ratio of the planner’s progress along the route to the expert’s progress. This ratio is clipped to the interval $[0,1]$.

Speed-limit Compliance (SC) assesses whether the planner adheres to the speed limit of the current lane. The resulting score $\mathrm{score}_{SC} \in [0,1]$ is 1 if there are no speed limit violations and approaches 0 as the violation increases.

Comfort (C) verifies whether the ego vehicle’s kinematic quantities, such as acceleration and jerk, remain within predefined thresholds. If all thresholds are satisfied, the metric returns $\mathrm{score}_{C} = 1$, otherwise $\mathrm{score}_{C} = 0$.

The final trajectory score combines all metrics according to:
\begin{equation}
\label{eq:nuplan_score}
\mathrm{Scenario Score} = \prod_{i \in \text{\tiny multiplier metrics}} \mathrm{Score}_i 
\times \sum_{j \in \text{\tiny weighted metrics}} w_j \cdot \mathrm{Score}_j
\end{equation}
\begin{table}[t]
\centering
\vspace{-0.2cm}
\begin{tabular}{lcc}
\toprule
\textbf{Metric} & \textbf{Weight} $w$ & \textbf{Range} \\
\midrule
No at-fault Collisions (NC) & multiplier & $\{0, \frac{1}{2},1\}$ \\
Drivable Area Compliance (DAC) & multiplier & $\{0,1\}$ \\
Driving Direction Compliance (DDC) & multiplier & $\{0,\frac{1}{2},1\}$ \\
Making Progress (MP) & multiplier & $\{0,1\}$ \\
\midrule
Time to Collision (TTC) within bound & 5 & $\{0,1\}$ \\
Ego Progress (EP) & 5 & $[0,1]$ \\
Speed-limit Compliance (SC) & 4 & $[0,1]$ \\
Comfort (C) & 2 & $\{0,1\}$ \\
\bottomrule
\end{tabular}
\caption{Summarizes the metrics and corresponding weights used for evaluating planning performance. The first group of metrics: No at-fault Collisions (NC), Drivable Area Compliance (DAC), Driving Direction Compliance (DDC), and Making Progress (MP), act as hard multipliers that can nullify a score if violated. The second group includes soft metrics weighted by importance: Time-to-Collision (TTC), Ego Progress (EP), Speed-limit Compliance (SC), and Comfort (C). Each metric’s valid range is provided, indicating whether it is binary, discrete, or continuous. These weights define how individual components contribute to the overall composite planning score.}
\vspace{-0.2cm}
\label{tab:suppl_metric_weights}
\end{table}
  
\subsection{Traffic Agents}
\label{subsec:suppl:traffic-agents}
As mentioned in Sec.~\ref{sec:benchmarks}, we use three different types of traffic agents in simulation:
\begin{itemize}
    \item \textit{Non-reactive (NR)}: in this setting surrounding agents follow the pre-recorded trajectories available in the dataset. If the ego deviates from the expert trajectory, the surrounding agents are non-reactive to the ego.
    \item \textit{Reactive (R)}: For surrounding agents, the current lane is extracted from the map graph, and the Intelligent Driver Model (IDM)~\cite{treiber2000congested} then computes a longitudinal trajectory along the chosen centerline. Given the current longitudinal position $x$, velocity $v$, and the distance $s$ to the leading vehicle along the centerline, IDM iteratively applies the following policy to calculate the longitudinal acceleration:
\begin{equation}
\frac{dv}{dt} = a \left[ 1 - \left(\frac{v}{v_0}\right)^\delta - \left(\frac{s^*}{s}\right)^2 \right] \quad ,
\end{equation}
where the acceleration limit $a$, target speed $v_0$, safety margin $s^*$, and exponent $\delta$ are manually chosen. 
Intuitively, this policy accelerates a vehicle unless it is already near $v_0$ or the leading vehicle is within the safety distance $s^*$. This makes the surrounding agents reactive to the ego vehicle.
\item \textit{SMART-reactive (SR)}~\cite{hagedorn2025plannersmeetreality}: this recent work builds on the SMART model~\cite{wu2024smart} to generate the behavior of surrounding agents. SMART is an autoregressive generative model that employs a decoder-only transformer to predict motion tokens for multiple agents, conditioned on previous motion tokens, the interactive motion of other agents, and encoded road context and is trained to predict the next motion token. SMART has ranked first place on the Waymo Open Motion Dataset (WOMD) leaderboards. In~\cite{hagedorn2025plannersmeetreality}, SMART was re-trained on the nuPlan dataset using a action-token vocabulary with 1024 discrete motion and road tokens each. Because the model is trained on human driving data, it can reproduce more human-like behavior than rule-based agents such as IDM. Such realism in traffic agent behavior is crucial for developing stronger planners and for reliably evaluating existing models.
\end{itemize}

\end{document}